\documentclass{article}

\usepackage{microtype}
\usepackage{graphicx}
\usepackage{subfigure}
\usepackage{booktabs} %

\usepackage[accepted]{icml2025}

\usepackage{amsmath}
\usepackage{amssymb}
\usepackage{mathtools}
\usepackage{amsthm}

\theoremstyle{plain}

\theoremstyle{definition}

\theoremstyle{remark}

\usepackage[textsize=tiny]{todonotes}

\icmltitlerunning{Recommendations and Reporting Checklist for Rigorous \& Transparent Human Baselines in Model Evaluations}

\usepackage{tabularx}
\usepackage{comment}
\usepackage{enumitem}
\usepackage{xurl}
\usepackage{xcolor}
\usepackage{soul}
\usepackage[T1]{fontenc} %
\usepackage{multirow}
\usepackage{multicol}
\usepackage{makecell}
\usepackage{caption}

\newcolumntype{Y}{>{\centering\arraybackslash}X}

\usepackage{hyperref}

\usepackage[capitalize,noabbrev]{cleveref}

\hypersetup{
   breaklinks=true,   %
   linktocpage=true,    %
}

\begin{document}

\pagenumbering{roman}

\twocolumn[
\icmltitle{Recommendations and Reporting Checklist for Rigorous \& Transparent \\ Human Baselines in Model Evaluations}

\icmlsetsymbol{equal}{*}

\begin{icmlauthorlist}
    \icmlauthor{Kevin L. Wei}{equal,R,H}
    \icmlauthor{Patricia Paskov}{equal,R,I}
    \icmlauthor{Sunishchal Dev}{equal,R,AV}
    \icmlauthor{Michael J. Byun}{equal,R,I}
    \icmlauthor{Anka Reuel}{H,S}
    \icmlauthor{Xavier Roberts-Gaal}{H}
    \icmlauthor{Rachel Calcott}{H}
    \icmlauthor{Evie Coxon}{MPS}
    \icmlauthor{Chinmay Deshpande}{CDT}
\end{icmlauthorlist}

\icmlaffiliation{H}{Harvard University, Cambridge, MA, USA}
\icmlaffiliation{I}{Independent}
\icmlaffiliation{AV}{Algoverse}
\icmlaffiliation{S}{Stanford University, Stanford, CA, USA}
\icmlaffiliation{CDT}{Center for Democracy \& Technology, Washington, D.C., USA}
\icmlaffiliation{MPS}{Max Planck School of Cognition, Leipzig, Germany}
\icmlaffiliation{R}{Technology \& Security Policy Fellow, RAND, Santa Monica, CA, USA. Views, opinions, findings, conclusions, \& recommendations contained herein are the authors' alone and not those of RAND or its research sponsors, clients, or grantors.}%

\icmlcorrespondingauthor{Kevin L. Wei}{kevinwei@acm.org}

\icmlkeywords{Human baseline, human performance baseline, AI evaluation, ML evaluation, evaluation methodology, science of evaluation}

\vskip 0.3in
]

\printAffiliationsAndNotice{\icmlEqualContribution} %

\section*{Executive Summary} \label{sec:Exec_Summary} \addcontentsline{toc}{section}{Executive Summary}

This paper finds that existing human baselines are neither sufficiently rigorous nor  transparent to enable meaningful comparisons of human vs. AI performance. We provide recommendations and a reporting checklist to increase rigor and transparency in human baselines. 

Human baselines are reference sets of metrics intended to represent human performance on specific tasks. They are used in AI evaluations to compare human vs. AI performance on evaluation items, adding important context to results and helping inform stakeholders in the broader AI ecosystem (e.g., downstream users, policymakers). 

Specifically, this paper makes three contributions:
\begin{enumerate}
    \item \textbf{Methodological recommendations}: Based on a meta-review of the measurement theory and AI evaluation literatures, we provide methodological recommendations for evaluators to build rigorous human baselines in AI evaluations. Recommendations are summarized in Figure \ref{fig:Exec_Summary_Framework}, with more details in Table \ref{tab:Exec_Summary_Recs}.
    \item \textbf{Reporting checklist}: We provide a reporting checklist for evaluators to increase transparency when publishing human baselines. The full checklist is in Appendix \ref{sec:Appendix_Checklist}.
    \item \textbf{Literature review}: We review 115 human baselines (studies) to identify methodological gaps in existing AI evaluations, and we find substantial shortcomings in the rigor and transparency of existing human baselines. Summary statistics of our review are in Table \ref{tab:Exec_Summary_Stats}, with more statistics and figures in Appendix \ref{sec:Appendix_Statistics}.
\end{enumerate}

Maximal rigor may not be possible in all human baselines due to resource limitations. In these cases, we hope to help researchers make informed tradeoffs, discuss/acknowledge methodological limitations, narrow interpretation of results, and transparently report methods and results. 

Data, code, and Word/LaTeX versions of our recommendations and reporting checklist are available at: \url{https://github.com/kevinlwei/human-baselines}.

\section*{Readers' Guide\footnote{Inspired by \citet{weidinger_ethical_2021, weidinger_sociotechnical_2023,wei_how_2024}.}} \label{sec:Readers_Guide} \addcontentsline{toc}{section}{Readers' Guide}

We recommend the following reading strategies for different types of readers:

\begin{itemize}
    \item \textbf{2-minute read}: Read Figure \ref{fig:Exec_Summary_Framework} and Tables \ref{tab:Exec_Summary_Stages}--\ref{tab:Exec_Summary_Stats}.
    \item \textbf{10-minute read}: Read Figure \ref{fig:Exec_Summary_Framework} and Tables \ref{tab:Exec_Summary_Stages}--\ref{tab:Exec_Summary_Stats}. If needed, skim the relevant parts of Section \ref{sec:Framework} to understand the rationale behind specific recommendations.
    \item \textbf{Stakeholders seeking to assess the quality of human baselines but who are not building human baselines} (e.g., policymakers, AI governance researchers, AI researchers who don't work on evaluations): Read Figure \ref{fig:Exec_Summary_Framework} and Tables \ref{tab:Exec_Summary_Stages}--\ref{tab:Exec_Summary_Stats}. If needed, skim the relevant parts of Section \ref{sec:Framework} for clarity about the meaning of or the rationale behind specific recommendations. Read Appendix \ref{sec:Appendix_Case_Studies} for examples of better/worse human baselines.
    \item \textbf{AI evaluators building human baselines}: Start with Figure \ref{fig:Exec_Summary_Framework} and Tables \ref{tab:Exec_Summary_Stages}--\ref{tab:Exec_Summary_Recs}. Read Section \ref{sec:Framework} to understand the rationale behind our recommendations, and read Appendix \ref{sec:Appendix_Expert_Baselines} if considering an expert human baseline. Use the reporting checklist in Appendix \ref{sec:Appendix_Checklist} when writing up results. Optionally, read the discussion in Sections \ref{sec:Discussion} and \ref{sec:Alternative_Views}, and read the case studies in Appendix \ref{sec:Appendix_Case_Studies} for examples of better/worse human baselines.
\end{itemize}

\renewcommand{\thefigure}{\Roman{figure}}
\begin{figure*}[!htbp]
    \centering
    \includegraphics[width=0.95\textwidth]{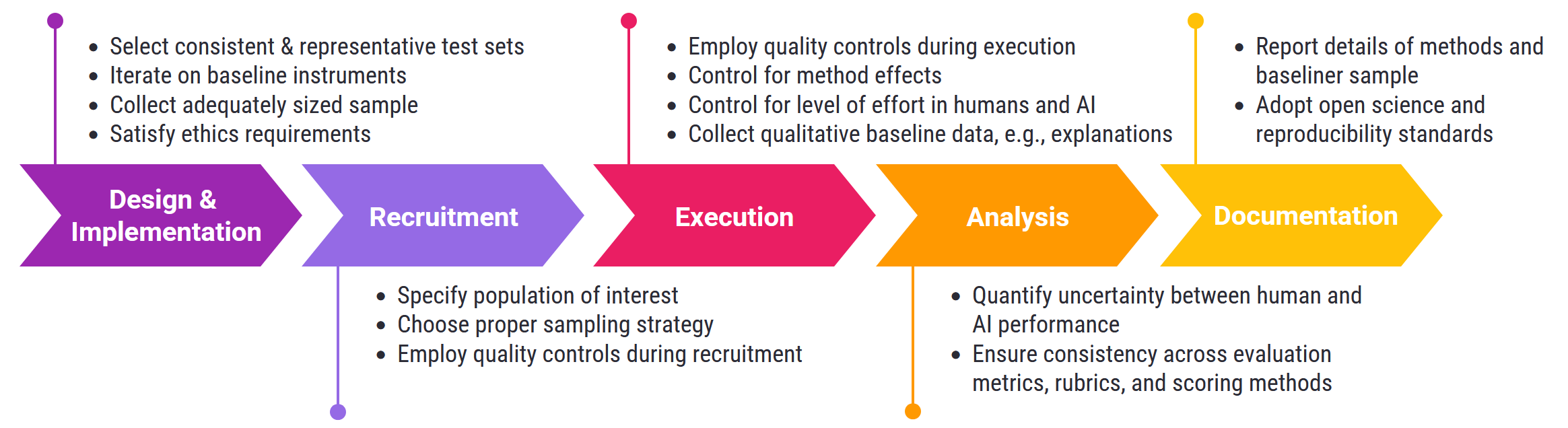}
    \caption{A summary of our recommendations for robust and transparent human baselines. Definitions of each stage of the baseline lifecycle are provided in Table \ref{tab:Exec_Summary_Stages}, and more details about our recommendations are provided in Table \ref{tab:Exec_Summary_Recs}. Full recommendations are in Section \ref{sec:Framework} and full checklist is in Appendix \ref{sec:Appendix_Checklist}.}
    \label{fig:Exec_Summary_Framework}
\end{figure*}

\renewcommand{\thefigure}{\arabic{figure}}
\setcounter{figure}{0}

\renewcommand{\thetable}{\Roman{table}}
\begin{table*}[!htbp]
    \centering
    \small
    
    \begin{tabularx}{\textwidth}{p{0.3\textwidth} X }
        \toprule
        
        \textbf{Human Baseline Stage} &
        \textbf{Definition} %
        \\

        \midrule

        Baseline Design \& Implementation &
        Baseline design is the initial stage of human baseline development, at which researchers define baselines’ purpose, scope, concepts, evaluation items, and metrics; baseline implementation is the selection and construction of tools and datasets for evaluation.
        \\

        Baseliner Recruitment &
        Baseliner recruitment is the stage at which human baseliners---the humans who respond to evaluation items---are found and are engaged to participate in a baseline.
        \\

        Baseline Execution & 
        Baseline execution is the stage at which the human baseline is conducted and result data is collected---e.g., through surveys or crowdwork platforms.
        \\

        Baseline Analysis & 
        Baseline analysis is the stage after data collection at which human baseline data is inspected and compared to AI results.
        \\

        Baseline Documentation & 
        Baseline documentation is the provision of evaluation tasks, datasets, metrics, and experimental materials and resources to relevant audiences.
        \\
        
        \bottomrule
    \end{tabularx}
    \caption{Definition of different stages in the human baseline lifecycle.}
    \label{tab:Exec_Summary_Stages}
\end{table*}

\begin{table*}[!t]
    \centering
    \small
    
    \begin{tabularx}{\textwidth}{p{0.3\textwidth} X }
        \toprule
        
        \textbf{Recommendation} &
        \textbf{Details} %
        \\

        \midrule

        \multicolumn{2}{l}{\textbf{Baseline Design \& Implementation}}%
        \\

        Use consistent \& representative test sets for human baselines and AI results &
        \begin{itemize}[topsep=0pt]
            \vspace{-0.75em}
            \item Use the same test set for human baselines and AI results.
            \item If using a subset of the full test set for human baselines, calculate AI results on that subset and make comparisons only on that subset.
            \item If using a subset of the full test set for human baselines, select the subset randomly or stratify on relevant criteria (e.g., based on question difficulty, topic, dataset source, etc.).
        \end{itemize} %
        \\

        Iteratively develop baseline instruments &
        \begin{itemize}[topsep=0pt]
            \vspace{-0.75em}
            \item Validate, collect feedback, and refine human baseline instruments such as survey questions, instructions, training materials (analogous to refining AI prompts or other materials).
            \item Examples of iterative development processes include (roughly in ascending order of cost/effort): expert validation, pre-tests of baseline instruments, focus groups, and pilot studies.
        \end{itemize}
        \\

        Collect an adequately sized sample of baseliners &
        \begin{itemize}[topsep=0pt]
            \vspace{-0.75em}
            \item For generalist (non-expert) baselines, conduct a statistical power analysis to ensure that your baseliner sample size is sufficiently large to represent the human population of interest for your baseline.
            \item A rule of thumb is that a sample of 1,000 respondents is needed to represent the U.S. adult population
            \item If baseliner samples must be smaller due to resource limitations, consider: 1) narrowing population of interest, 2) calculating and reporting the required sample size to reliably detect effects (even if researchers are unable to collect a sample of that size), and 3) narrowing interpretations of baseline results.
            \item For expert baselines, see discussion in Appendix \ref{sec:Appendix_Expert_Baselines}.
        \end{itemize}
        \\

        Satisfy ethics requirements for human subjects research &
        \begin{itemize}[topsep=0pt]
            \vspace{-0.75em}
            \item Ensure ethics requirements are followed, e.g., collecting informed consent, ethics (IRB) review, and other human subjects protections.
            \item Report which ethics requirements were necessary/satisfied for your baseline.
        \end{itemize}
        \\

        \midrule

        \multicolumn{2}{l}{\textbf{Baseliner Recruitment}}
        \\

        Specify a human population of interest &
        \begin{itemize}[topsep=0pt]
            \vspace{-0.75em}
            \item Specify which subset of humans the baseline is intended to represent.
            \item Define your population using dimensions such as geographic location, demographic characteristics (e.g., age, gender, socioeconomic status), language, cultural background, education, or domain expertise.
            \item Narrow your population of interest so that you aren't attempting to estimate metrics for all humans.
        \end{itemize}
        \\

        Use an appropriate sampling strategy for selecting baseliners &
        \begin{itemize}[topsep=0pt]
            \vspace{-0.75em}
            \item For generalist (non-expert) baselines, random sampling is ideal, but crowdwork samples are often the norm due to cost considerations. When using crowdwork samples, consider methodological adjustments to make your sample more representative of the underlying human population of interest---e.g., stratified sampling, survey weights, the ``representative sample'' option in Prolific, etc.
            \item For expert baselines, convenience samples are acceptable. Clearly define criteria for baseliner eligibility, and consider snowball sampling. See Appendix \ref{sec:Appendix_Expert_Baselines} for additional discussion and case studies in Appendix \ref{sec:Appendix_Case_Studies}.
        \end{itemize}
        \\

        \bottomrule
    \end{tabularx}
    \caption{Methodological recommendations for rigorous human baselines (cont'd on next page)}
    \label{tab:Exec_Summary_Recs}
\end{table*}

\begin{table*}[!p]
    \centering
    \small
    
    \begin{tabularx}{\textwidth}{ p{0.3\textwidth} X }
        \toprule
        
        \textbf{Recommendation} &
        \textbf{Details} %
        \\

        \midrule

        \multicolumn{2}{l}{\textbf{Baseliner Recruitment (Cont'd)}}
        \\
        
        Employ quality controls for baseliner recruitment &
        \begin{itemize}[topsep=0pt]
            \vspace{-0.75em}
            \item Use inclusion/exclusion criteria for baseliners to ensure data quality.
            \item Consider using pre-qualification tests or screening questions, quality scores (on crowdwork platforms), and excluding baseliners previously exposed to evaluation items.
            \item Consider excluding the authors or other members of the research team as baseliners if they have been previously exposed to evaluation items (or would otherwise bias results).
        \end{itemize}
        \\

        \midrule

        \multicolumn{2}{l}{\textbf{Baseline Execution}}
        \\

        Employ quality controls during baseline execution &
        \begin{itemize}[topsep=0pt]
            \vspace{-0.75em}
            \item Exclude unreliable baseliner responses to improve data quality---e.g., filtering baseliner responses by outliers, time to completion, or other paradata.
            \item Consider including attention checks, comprehension/manipulation checks, consistency checks, or honeypot questions during the baseline itself.
            \item If baseliners aren't supposed to use AI tools during baselining, consider including instructions asking participants not to use AI tools, employing technical restrictions such as preventing copy/pasting, using non-standard interface elements, or using comprehension/manipulation checks.
        \end{itemize}
        \\

        Control for method effects and use identical tasks &
        \begin{itemize}[topsep=0pt]
            \vspace{-0.75em}
            \item Method effects are variations in item response attributable to data collection methods rather than to differences in underlying response distributions (e.g., due to instructions, option order, or mode of data collection).
            \item Use the same tasks for both human and AI responses (i.e., identical instructions, examples, context, etc.).
            \item Randomize question order, response option order, and other non-critical methodological details.
            \item Some method effects may be inevitable due to differences between human and AI cognition; consider clearly documenting methodological details and discussing these limitations. See additional discussion in Section \ref{subsec:Framework_Execution}.
        \end{itemize}
        \\

        Control for level of effort &
        \begin{itemize}[topsep=0pt]
            \vspace{-0.75em}
            \item Make fair comparisons by comparing human and AI results at similar levels of effort---e.g., when given the same amount of time to solve a task, or at similar levels of financial cost.
            \item AI effort can be affected by inference cost, task time limits, sampling or elicitation strategy, and other factors.
            \item Human baseliner effort can be affected by training, compensation, task time limits, and other factors.
        \end{itemize}
        \\

        Collect qualitative data from baseliners &
        \begin{itemize}[topsep=0pt]
            \vspace{-0.75em}
            \item Collect qualitative data from baseliners to help surface new insights, e.g., failure modes.
            \item Qualitative data can include baseliner explanations (about why they chose particular responses), task trajectories, etc.
            \item Qualitative data may not be necessary for all baselines, especially since collecting such data may increase the cost of baselines.
        \end{itemize}
        \\

        \bottomrule
    \end{tabularx}
    \caption*{Methodological recommendations for rigorous human baselines (Table \ref{tab:Exec_Summary_Recs}, cont'd)}
\end{table*}

\begin{table*}[!p]
    \centering
    \small
    
    \begin{tabularx}{\textwidth}{ p{0.3\textwidth} X }
        \toprule
        
        \textbf{Recommendation} &
        \textbf{Details} %
        \\

        \midrule
        
        \multicolumn{2}{l}{\textbf{Baseline Analysis}}
        \\

        Quantify uncertainty in human vs. AI performance differences &
        \begin{itemize}[topsep=0pt]
            \vspace{-0.75em}
            \item Report measurements of uncertainty rather than just point estimates---e.g., statistical tests, interval estimates, or distributions of performance.
        \end{itemize}
        \\

        Use consistent evaluation metrics, scoring methods, and rubrics across human and AI evaluation &
        \begin{itemize}[topsep=0pt]
            \vspace{-0.75em}
            \item Use the same evaluation metrics, scoring methods, and scoring rubrics for both human and AI results.
            \item When using aggregate metrics such as pass@k, majority vote, etc., consider using the same aggregate metrics for both human and AI performance (or explaining why it may be appropriate to use different metrics).
        \end{itemize}
        \\

        \midrule

        \multicolumn{2}{l}{\textbf{Baseline Documentation}}
        \\

        Report key details about baselining methodology and baseliners &
        \begin{itemize}[topsep=0pt]
            \vspace{-0.75em}
            \item Report information about baseliners, baselining procedures, and baseline paradata. These details are important for interpreting and assessing baseline results.
            \item Consider using the reporting checklist we provide in Appendix \ref{sec:Appendix_Checklist}.
        \end{itemize}
        \\

        Adopt best practices for open science and reproducibility/replicability &
        \begin{itemize}[topsep=0pt]
            \vspace{-0.75em}
            \item Where possible, release (anonymized) human baseline data, experimental materials such as forms or custom UIs, and analysis code.
            \item Releasing such data helps validate research and may promote re-use of your human baseline in future work.
        \end{itemize}
        \\         

        \bottomrule
    \end{tabularx}
    \caption*{Methodological recommendations for rigorous human baselines (Table \ref{tab:Exec_Summary_Recs}, cont'd)}
\end{table*}

\begin{table*}[!p]
    \centering
    \small
    \begin{tabularx}{\textwidth}{p{0.4\textwidth} YYY YYY }
        \toprule
        \multirow{2}{*}{\textbf{Question}} & \multicolumn{3}{c}{\textbf{All Baselines} ($n = 115$)} & \multicolumn{3}{c}{\textbf{Model Card Baselines} ($n = 7$)} \\
         & Yes & No & Unknown
         & Yes & No & Unknown
         \\
         
        \midrule
        \multicolumn{5}{l}{\textbf{Baseline Design \& Implementation}}
        \\

        \textbf{Test Set Equivalence}: Were human and AI test sets identical? (Default: No) & 
            \makecell[t]{59.13\% \\ 68} & 
            \makecell[t]{40.87\% \\ 47} &
            &
            \makecell[t]{57.14\% \\ 4} &
            \makecell[t]{42.86\% \\ 3} &
            \\

        1.6 \textbf{Iterative Design}: Was the experimental setup of the baseline iteratively designed with participatory methods? & 
            \makecell[t]{34.78\% \\ 40} &
            \makecell[t]{12.17\% \\ 14} &
            \makecell[t]{63.04\% \\ 61} &
            \makecell[t]{42.86\% \\ 3} &
            \makecell[t]{28.57\% \\ 2} &
            \makecell[t]{28.57\% \\ 2} 
            \\
        
        1.7 \textbf{Amount of Effort}: Does the baseline control for the amount of effort by human baseliners and AIs & 
            \makecell[t]{13.91\% \\ 16} & 
            \makecell[t]{35.65\% \\ 41} &
            \makecell[t]{50.43\% \\ 58} & 
            \makecell[t]{28.57\% \\ 2} &
            \makecell[t]{57.14\% \\ 4} &
            \makecell[t]{14.29 \\ 1} 
            \\

        1.8 \textbf{Power Analysis}: Did the authors conduct power analysis in order to determine baseline size? (Default: No) & 
            \makecell[t]{1.74\% \\ 2} & 
            \makecell[t]{98.26\% \\ 113} &
            &
            \makecell[t]{0.00\% \\ 0} &
            \makecell[t]{100.00\% \\ 7} &
            \\

        1.9 \textbf{Ethics Review}: Was the study approved or exempted by an IRB, or did it undergo other ethics review?  & 
            \makecell[t]{13.91\% \\ 16} & 
            \makecell[t]{2.61\% \\ 3} &
            \makecell[t]{83.48\% \\ 96} & 
            \makecell[t]{0.00\% \\ 0} &
            \makecell[t]{14.29\% \\ 1} &
            \makecell[t]{85.71\% \\ 6} 
            \\

        \midrule

        \multicolumn{5}{l}{\textbf{Baseliner Recruitment}}
        \\

        2.1 \textbf{Population of Interest Identification}: Does the reporting identify human populations for which these results may be valid, i.e., a human population of interest? (Default: No) & 
            \makecell[t]{42.61\% \\ 49} & 
            \makecell[t]{57.39\% \\ 66} &
            &
            \makecell[t]{57.14\% \\ 4} &
            \makecell[t]{42.86\% \\ 3} &
            \\

        2.3 \textbf{Quality Control in Recruitment}: Were human baseliners pre-qualified or excluded during the recruitment process for any reason? (Default: Yes) & 
            \makecell[t]{28.70\% \\ 33} & 
            \makecell[t]{71.30\% \\ 82} &
            &
            \makecell[t]{28.57\% \\ 2} &
            \makecell[t]{71.43\% \\ 5} &
            \\

        \midrule

        \multicolumn{5}{l}{\textbf{Baseline Execution}}
        \\

        3.2 \textbf{Quality Control in Execution}: Were quality checks implemented or data cleaned/excluded during the data collection process (i.e., after baseliners were recruited)? (Default: No) & 
            \makecell[t]{23.48\% \\ 27} & 
            \makecell[t]{76.52\% \\ 88} &
            &
            \makecell[t]{28.57\% \\ 2} &
            \makecell[t]{71.43\% \\ 5} &
            \\

        3.4 \textbf{Instruction Equivalence}: Did the human baseliners and AIs have access to the same instructions/prompt/question for each item? (Default: No) & 
            \makecell[t]{24.35\% \\ 28} & 
            \makecell[t]{75.65\% \\ 87} &
            &
            \makecell[t]{14.29\% \\ 1} &
            \makecell[t]{85.71\% \\ 6} &
            \\

        3.5 \textbf{Tool Access Equivalence}: Did the human baseliners and AIs have access to the same (technical) tools for each item? (Default: Yes) & 
            \makecell[t]{89.57\% \\ 103} & 
            \makecell[t]{10.43\% \\ 12} &
            &
            \makecell[t]{71.43\% \\ 5} &
            \makecell[t]{28.57\% \\ 2} &
            \\

        3.6 \textbf{Explanations}: Did the eval/baseline collect explanations from the human baseliners, after the evaluation was conducted? (Default: No) & 
            \makecell[t]{11.30\% \\ 13} & 
            \makecell[t]{97.39\% \\ 112} &
            &
            \makecell[t]{42.86\% \\ 3} &
            \makecell[t]{57.14\% \\ 4} &
            \\

        \midrule
        
        \multicolumn{5}{l}{\textbf{Baseline Analysis}}
        \\

        4.1 \textbf{Statistical Significance}: Did the eval test for statistically significant differences between AI and human performance? (Default: No) & 
            \makecell[t]{8.70\% \\ 10} & 
            \makecell[t]{91.30\% \\ 105} &
            &
            \makecell[t]{0.00\% \\ 0} &
            \makecell[t]{100.00\% \\ 7} &
            \\

        4.2 \textbf{Uncertainty Estimate}: Did the paper present a measure of uncertainty for the AI and human baseline results? (Default: No) & 
            \makecell[t]{33.04\% \\ 38} & 
            \makecell[t]{66.96\% \\ 77} &
            &
            \makecell[t]{14.29\% \\ 1} &
            \makecell[t]{85.71\% \\ 7} &
            \\

        4.3 \textbf{Evaluation Metric Equivalence}: Was the same evaluation metric measured/compared for both humans and AIs? (Default: Yes) & 
            \makecell[t]{93.91\% \\ 108} & 
            \makecell[t]{6.09\% \\ 7} &
            &
            \makecell[t]{100.00\% \\ 7} &
            \makecell[t]{0.00\% \\ 0} &
            \\

        \midrule

        \multicolumn{5}{l}{\textbf{Baseline Documentation}}
        \\
        5.1.1 \textbf{Reporting Sample Demographics}: Were demographics for human baseliners, e.g., race, gender, etc. reported? (Default: No) & 
            \makecell[t]{22.61\% \\ 26} & 
            \makecell[t]{77.39\% \\ 89} &
            &
            \makecell[t]{28.57\% \\ 2} &
            \makecell[t]{71.43\% \\ 5} &
            \\

        5.2 \textbf{Baseline Data Availability}: Is the (anonymized) human baseline data publicly available? (Default: No) & 
            \makecell[t]{21.74  \% \\ 25} & 
            \makecell[t]{78.26\% \\ 90} &
            &
            \makecell[t]{28.57\% \\ 2} &
            \makecell[t]{71.43\% \\ 5} &
            \\

        \bottomrule
    \end{tabularx}
    \caption{Summary statistics from our literature review of 115 existing human baselines, with the same results on a subset of 7 evaluations commonly used in industry model cards (\textsc{MMMU}, \textsc{GPQA}, \textsc{MATH}, \textsc{DROP}, \textsc{ARC}, \textsc{ConceptARC}, and \textsc{EgoSchema}). ``Unknown'' values mean that the item was not reported in text; rows without unknowns are imputed to default values (specified per-question above). More figures and statistics are in Appendix \ref{sec:Appendix_Statistics}.}
    \label{tab:Exec_Summary_Stats}
\end{table*}

\renewcommand{\thetable}{\arabic{table}}
\setcounter{table}{0}

\clearpage

\onecolumn
\makeatletter
\setcounter{tocdepth}{2}
\tableofcontents \addcontentsline{toc}{section}{Table of Contents}

\clearpage

\listoffigures \addcontentsline{toc}{section}{List of Figures}
\listoftables \addcontentsline{toc}{section}{List of Tables}

\clearpage

\twocolumn

\pagenumbering{arabic}

\section{Introduction}
\label{sec:Intro}

Artificial intelligence (AI) systems, foundation models in particular, have increasingly achieved superior performance on benchmarks in natural language understanding, general reasoning, coding, and other domains \cite{maslej_chapter_2024}. These results are frequently compared to \textit{human baselines}---reference sets of metrics intended to represent human performance on specific tasks---which has led to claims about models' ``super-human'' performance \cite{bikkasani_navigating_2024}. 

Human baselines are crucial for evaluating AI systems and for understanding AI's societal impacts. For the machine learning (ML) research community, human baselines help improve benchmarks, provide context for interpreting system performance, and demonstrate concurrent validity \cite{hardy_more_2024, bowman_what_2021}. For downstream users, comparisons to human performance may inform decisions about AI adoption \citetext{cf. \citealt{luo_frontiers_2019}}. And for policymakers, human baselines facilitate risk assessments \cite{ostp_blueprint_2022, nist_ai_2023, goemans_safety_2024, us_aisi_us_2024} and predictions of AI's economic impacts \cite{hatzius_potentially_2023, shrier_is_2023}. Valid and reliable human baselines thus contribute greatly to the operational value of AI evaluations.

However, despite widespread recognition in the ML community about the importance of human baselines \cite{reuel_betterbench_2024, ibrahim_beyond_2024, tedeschi_whats_2023, nangia_human_2019, bender_establishing_2015}, existing human baselines used currently to assess human performance on a wide array of AI evaluation tasks (including reasoning, coding, visual perception, etc.) are neither sufficiently rigorous nor sufficiently transparent to enable reliable claims about (the magnitude of) differences between human and AI performance. For instance, human baselines in many evaluations have small or biased samples \cite{liao_are_2021, mcintosh_inadequacies_2024}, apply different instruments than those used in AI evaluation \cite{tedeschi_whats_2023}, or fail to control for confounding variables \cite{cowley_framework_2022}. In addition, published evaluations commonly omit study details necessary for assessing baseline validity, such as how participants were recruited or how questions were administered (Section \ref{subsec:Framework_Documentation}). Measurement theory, a methodological field in the social sciences concerned with quantifying complex concepts, addresses analogous issues in human studies \cite{bandalos_measurement_2018} and can inform best practices in human baselines.

\textbf{Our position is that human baselines in evaluations of foundation models must be more rigorous and more transparent.} Building from measurement theory, we propose recommendations for producing more rigorous human baselines. We also synthesize our recommendations into a reporting checklist, which we use to systematically review 115 published human baselines, finding substantial shortcomings in existing human baselining methods. We hope that our recommendations and reporting checklist can support researchers in developing and documenting human baselines that are more interpretable and valuable to the ML community, downstream users, and policymakers. 

In defending our position, we believe evaluators should be expected to conduct baselines with significant rigor to enable performance comparisons. However, we acknowledge that there are often barriers to rigor, including the expense of high-quality baseline data, the evolving evaluation landscape, and differences in cognition and interaction modes between humans and AI systems. Where maximal rigor is infeasible, evaluators should discuss limitations and narrow their interpretations of baseline comparisons. Our work highlights some of these limitations where applicable and aims to support evaluators in making conscious decisions about tradeoffs between experimental rigor and practical considerations such as cost and efficiency.

We proceed to discuss background in Section \ref{sec:Background}. Section \ref{sec:Methods} describes our methodology (details in Appendix \ref{sec:Appendix_Methods}). Section \ref{sec:Framework} presents our recommendations (full reporting checklist in Appendix \ref{sec:Appendix_Checklist}) and results of our systematic review, which examines the entire lifecycle of human baselines: baseline(r) design, recruitment, execution, analysis, and documentation. Section \ref{sec:Discussion} contains discussion and limitations, and Section \ref{sec:Alternative_Views} surveys alternative views. Section \ref{sec:Conclusion} concludes. %

\section{Background}
\label{sec:Background}

Measurement theory is the discipline devoted to quantifying complex or unobservable concepts through the use of observable indicators, or measurements \cite{goertz_social_2020}. Concepts are often multidimensional or impossible to measure directly, so researchers usually aggregate multiple measurements and rely on proxies for quantities of interest. Intelligence, for instance, has sometimes been measured by aggregating multiple different cognitive tests \cite{deary_intelligence_2012}. In the social sciences, measurement theory has also been applied to concepts like fairness \cite{patty_measuring_2019}, emotion \cite{reisenzein_measuring_2024}, culture \cite{mohr_problems_2014}, personality \cite{drasgow_test_2009}, and language \cite{sassoon_measurement_2010}. Measurement theory helps build indicators for these concepts that satisfy criteria of validity (yielding results that support intended interpretations of measurements) and reliability (yielding consistent results across many measurements) \cite{bandalos_measurement_2018, salaudeen_measurement_2025}.%

There has been growing recognition in the AI research community that AI evaluation can learn from measurement theory and the social sciences \cite{chang_survey_2024, wallach_evaluating_2024, eckman_position_2025, chouldechova_shared_2024, blodgett_human-centered_2024, xiao_evaluating_2023, zhou_deconstructing_2022, zhao_position_2025, wang_evaluating_2023, liao_rethinking_2023, saxon_benchmarks_2024}. Like measurement theory, AI evaluation has been concerned with estimating concepts such as intelligence, fairness, emotion, and culture---though in AI models rather than in humans \cite{chang_survey_2024}. Recent research in ML has focused in particular on applying measurement theory to performance metrics \cite{subramonian_it_2023, flach_performance_2019} and fairness metrics \cite{jacobs_meaning_2020, grote_fairness_2024, blodgett_stereotyping_2021}. Additionally, measurement theory provides frameworks for making comparisons between (human) populations---analogous to the problem of comparing human and AI performance, which is often addressed using human baselines in AI evaluations. %

We draw on measurement theory to examine human baselining in evaluations of \textit{foundation models} \cite{bommasani_opportunities_2022}, which pose unique evaluation challenges \cite{liao_rethinking_2023}. Applying measurement theory to the foundation model context is particularly appropriate as foundation models are exhibiting increasingly general, multidimensional capabilities \cite{zhong_agieval_2024} and beginning to interact with the same interfaces as human users \cite{anthropic_developing_2024, chan_infrastructure_2025}. Specifically, we adopt the approach of \citet{zhao_position_2025} in drawing on measurement theory to validate the data generation process in human baselines---that is, we are particularly concerned with the validity and reliability of baselining \textit{methods}. 

Analysis of the full pipeline of human baselining methods in the foundation model context is limited. \citet{cowley_framework_2022} adapts best practices from psychology to human baselines in computer vision studies but does not examine the entire baseline lifecycle or provide operational-level recommendations. \citet{tedeschi_whats_2023} critiques baseline practices as part of a larger commentary on LLMs but does not examine related disciplines. And research in human-computer interaction methods has similarly been concerned with concept measurement \cite{lazar_research_2017}, though rarely with human baselines in particular. Building on this literature, we draw on measurement theory across the social sciences to both conceptual and operational methodological recommendations for human baselining in the context of foundation models. We also fill a gap in the literature by systematically reviewing human baselines in foundation model evaluations, allowing us to identify shortcomings of and opportunities for improvement in existing methods for human baselining.

\section{Methodology}
\label{sec:Methods}

We used a two-stage approach to develop our position, adapted from \citet{zhao_position_2025} and \citet{reuel_betterbench_2024}. First, we conducted a meta-review (review of reviews) of the measurement theory literature to construct best practices for baselining (Appendix \ref{sec:Appendix_Checklist}). Using purposive sampling and backwards snowballing, we identified 29 articles from the social sciences (psychology, economics, political science, education) and AI evaluation. We synthesized these recommendations into a more detailed reporting checklist, which was initially compiled after reviewing these articles and later refined through internal discussion and expert validation.

Second, using our reporting checklist, we conducted a systematic review \citetext{see \citealt{page_prisma_2021}} of the AI evaluations literature to identify gaps in existing human baselining methods. From academic publications and gray literature, we identified 115 human baselines in foundation model evaluations. Inclusion criteria consisted of whether the article contained 1) an original human baseline, 2) an evaluation of a foundation model, and 3) both a human baseline-related keyword (``human baseline*'', ``expert baseline*'', ``human performance baseline*'') and an AI evaluation-related keyword (``AI evaluation*'', ``ML benchmark*'', etc.). ``Baselines'' from observational data were excluded. Articles were then manually coded and validated per the checklist from our meta-review (Appendix \ref{sec:Appendix_Checklist}), with the codebook iteratively refined during the coding process. Full methodological details are in Appendix \ref{sec:Appendix_Methods}. 

\begin{figure*}[!thp]
    \centering
    \includegraphics[width=0.95\textwidth]{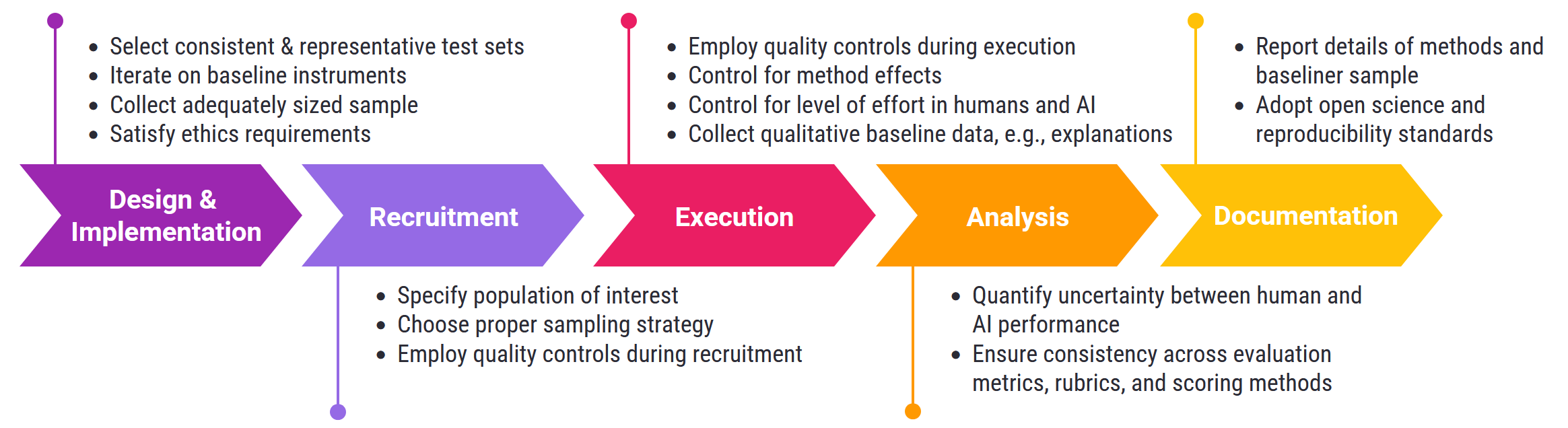}
    \caption{A summary of our recommendations for robust and transparent human baselines. Full recommendations in Section \ref{sec:Framework} and full checklist in Appendix \ref{sec:Appendix_Checklist}.}
    \label{fig:Framework}
\end{figure*}

\section{A Framework for Rigorous and Transparent Human Baselines}
\label{sec:Framework}

In this section, we provide high-level recommendations for conducting human baselines. We organize our discussion by delineating five stages of the baselining process, as adapted from \citet{reuel_betterbench_2024} and \citet{paskov_preliminary_2025}: design, recruitment, execution, analysis, and documentation. We summarize these stages and recommendations in Figure \ref{fig:Framework}, and we examine both positive and negative examples of human baseline studies in Appendix \ref{sec:Appendix_Case_Studies}.

We also discuss results of our systematic review. Appendix \ref{sec:Appendix_Checklist} has the full reporting checklist used in our review, and Appendix \ref{sec:Appendix_Statistics} contains select per-question summary statistics. Appendix \ref{sec:Appendix_Resources} has additional resources and practical guidance.

\subsection{Baseline Design \& Implementation}
\label{subsec:Framework_Design}

Baseline design is the initial stage of human baseline development, at which researchers define baselines' purpose, scope, concepts, evaluation items, and metrics; baseline implementation is the selection and construction of tools and datasets for evaluation \cite{reuel_betterbench_2024, paskov_preliminary_2025}. We examine four considerations for this stage. %

\textbf{Use consistent \& representative test sets for human baselines and AI results.} Robust comparisons of human vs. AI performance require comparing performance on the same test set. Where the human baseline's test set is a subset of the AI test set, performance comparisons should only be made on the subset, and the subset should also be representative of the underlying set. Of the baselines in our review, 41\% used different test sets for AI vs. human baselines. 

Because the cost of human baselines can make baselining on a large dataset infeasible, researchers often construct baselines using subsets of the test sets used for AI evaluation. Baseline validity thus depends on the sampling strategy used to create the human baseline test set. Simple random sampling from the broader evaluation dataset may be sufficient to ensure representativeness of the baseline test set when the test set is sufficiently large \citetext{see \citealt{liao_are_2021}}. Stratified sampling may be preferred where the baseline test set is relatively small, or where the test set must preserve important properties of the evaluation dataset such as data source \citetext{e.g., \citealt{xiang_care-mi_2023}}, question difficulty \cite{tedeschi_whats_2023}, or other relevant dimensions \citetext{\citealt{cowley_framework_2022}; e.g., \citealt{liu_tempcompass_2024}, \citealt{bai_power_2024}; see also \citealt{siska_examining_2024}}. Baseline test sets (where specified and distinct from AI test sets) were most commonly created using simple random (43\%), stratified (38\%), or purposeful sampling strategies (6\%). 

In addition, researchers should clearly indicate where human baseline test sets differ from AI evaluation test sets when reporting human baseline results. To directly compare AI results with human baselines, researchers should also report AI performance on only the human baseline test set.

\textbf{Iteratively develop baseline instruments.} Iterative processes repeatedly test and refine the measurement instruments (e.g., forms, surveys) by applying multiple rounds of validation, feedback, and refinement of before final data collection. Although researchers often iterate on prompts for AI systems, only 35\% of baselines reviewed reported iteratively developing human baseline instruments.

The feedback loops created by iteration can support construct validity \cite{rosellini_developing_2021} while ensuring clarity and consistent interpretation of instruments \cite{cheng_human_2024, cowley_framework_2022}. In the social sciences, iterative processes are the gold standard for collecting annotations \cite{cheng_human_2024}, running surveys \cite{groves_survey_2011}, and building clinical questionnaires \cite{rosellini_developing_2021}. The ML community has also recognized the importance of iteration, such as when optimizing AI prompts \cite{hewing_prompt_2024, gao_prompt_2025}; researchers also often validate items in evaluation \textit{datasets} \citetext{e.g., \citealt{nangia_what_2021, rein_gpqa_2024}} but less frequently validate baseline \textit{instruments}. An evaluation that optimizes AI prompts and validates evaluation items, but that does not validate baselining instruments, may unfairly disadvantage humans and thereby discount baseline validity.

Iteration does add complexity to the baselining process. However, it is not necessarily costly: large pilot studies and focus groups may be out of reach to budget-constrained researchers, but small-scale pre-tests or expert validation could still improve measurement instruments \cite{groves_survey_2011, zickar_measurement_2020}. 

\textbf{Collect an adequately sized sample of baseliners.}\footnote{``Sample'' in this context refers to the subset of humans in the baseline who are drawn from an underlying population.} Baselines that are underpowered because of small sample sizes are unreliable because they cannot robustly capture the underlying distribution of human performance across a population \cite{cao_why_2024}. Power analyses can help determine an appropriate sample size for human baselines, given significance levels and pre-specified minimum detectable effect sizes in the outcome metric of interest \cite{mcnulty_power_2021, cohen_statistical_2013}. %
The importance of statistical power in ML benchmarks has been noted in prior work \cite{card_little_2020, bowman_what_2021, grosse-holz_early_2024, beyer_llm-safety_2025}, but only 2\% of the human baselines we reviewed reported conducting power analyses. A rule of thumb is that a sample size of 1,000 is needed to represent the population of U.S. adults with a reasonable margin of error \cite{gelman_how_2004}. By this standard, baselines are vastly underpowered since the median sample size in our review was 8 (mean 90, though with high variance).%

If sample sizes are fixed (e.g., due to cost constraints), researchers can nevertheless calculate and report the required sample size to reliably detect practically important effects, which supports interpretation of evaluation results. Understanding the ability of human baselines to detect performance differences may be especially important to users and policymakers, who may demand added rigor and certainty in evaluation results to inform decision-making \cite{paskov_gpai_2024}. In general, considerations around statistical power reflect broader shortcomings in using statistical methods in AI evaluation, which we discuss further in Section \ref{subsec:Framework_Analysis}.

\textbf{Satisfy ethics requirements for human subjects research.} Ethics requirements---such as ethics review and collecting informed consent---protect human research participants \cite{page_improving_2017}; ethics review is legally required in many jurisdictions, including the U.S. \cite{us_department_of_homeland_security_45_2017}. Significantly, only 14\% of the articles we examined reported compliance with or formal exemption from ethics review requirements, near the same order of magnitude as the 2\% found by \citet{kaushik_resolving_2024} \citetext{see also \citealt{mckee_human_2024}}.

Reporting compliance with ethics requirements is best practice in many fields, e.g., medicine \cite{icmje_recommendations_2025}. Failure to report compliance in an article does not indicate failure to comply, and some evaluations may be exempt from review \cite{kaushik_resolving_2024}. However, transparency around research ethics becomes more important as public interest in AI increases, and protection of research participants can also be critical for evaluations implicating, e.g., deception, misinformation, and psychological impacts. %

\subsection{Baseliner Recruitment}
\label{subsec:Framework_Recruitment}

Baseliner recruitment is the stage at which human baseliners---the humans who respond to evaluation items---are found and are engaged to participate in a baseline. We examine three considerations for baseliner recruitment. %

\textbf{Specify a human population of interest.} Specifying a population of interest---the group of humans for whom a baseline is intended to be representative---is necessary to interpret \textit{which} group of humans a baseline represents. Defining the population is important since population sizes can affect sampling reliability and statistical power, and researchers can choose more targeted populations to save on sample size (though cost of targeting may also increase). Prior work has noted that AI evaluations rarely specify populations of interest \cite{subramonian_it_2023}, which is in line our review: only 43\% baselines explicitly or implicitly defined a population of interest along at least one axis (i.e., a population beyond ``humans,'' which is too large for most baselines to represent). %

Populations of interest can be specified through axes such as geographic location, demographic characteristics (e.g., age, gender, socioeconomic status), language, cultural background, education, or domain expertise. A human baseline may seek to measure the performance of, for instance, a population of medical or legal professionals \citetext{e.g., \citealt{blinov_rumedbench_2022, hijazi_arablegaleval_2024}}. Of baselines in our review, among AI evaluations that defined a population of interest, the most commonly reported characteristics were education (21\%), language (19\%), age (19\%), and expertise (18\%). How to scope the population of interest for any given baseline will depend on the evaluation's research questions, context, and intended use.

\textbf{Use an appropriate sampling strategy for selecting baseliners.} Sampling strategy--the methodology by which baseliners are selected--directly informs how representative the baseliner sample is of the population of interest. Representativeness is essential for external validity because it determines whether baseliners' results can be generalized to that underlying population \cite{findley_external_2021, stantcheva_how_2023, berinsky_measuring_2017, lohr_sampling_2022, valliant_practical_2018}. In our review, 31\% of human baselines used a convenience sample, 32\% recruited from crowdsourcing platforms, none used a random sample, and 37\% did not report sampling strategies. %

Random samples are ideal when baselines are meant to mirror broad human populations (e.g., generalist baselines), since other sampling strategies like convenience sampling are susceptible to significant biases that reduce generalizability \citetext{cf. \citealt{mihalcea_why_2024, diaz_what_2024, brown_when_2023}}. Most generalist baselines are conducted through crowdsourcing platforms such as Amazon Mechanical Turk (MTurk) or Prolific, which are not random samples and can pose challenges to representativeness. Crowdsourced samples could be demographically biased---MTurk workers tend to more educated, politically liberal, online, and younger than the general population \cite{sheehan_crowdsourcing_2018, shaw_online_2021, stantcheva_how_2023}---or biased due to expertise if crowdworkers have been exposed to extensive AI evaluation/training tasks.\footnote{Prolific samples are less well-studied than MTurk samples, so less is known about their representativeness.} Crowdsourced baselines may thus fail to represent performance of humans not of those demographics; even very large samples may be biased if insufficiently representative of the population of interest \cite{bradley_unrepresentative_2021}. In expert baselines, however, convenience sampling may be justified because expert populations can be very small (see discussion in Appendix \ref{sec:Appendix_Expert_Baselines}). %

When random sampling is infeasible, as is often the case due to cost, researchers designing human baselines can consider methodological adjustments to improve representativeness. For instance, stratified sampling can improve representativeness along specific dimensions \cite{groves_survey_2011}, and researchers building generalist baselines on Prolific can consider the ``representative sample'' option \cite{prolific_representative_2025}.\footnote{This option helps with representativeness \textit{as long as the axes on which Prolific samples are also those that define population of interest}. As of this writing, MTurk has no comparable option.} Post hoc adjustments such as weighting (which may require collecting baseliners' demographic information) or other sampling adjustments may also partially mitigate selection bias in non-representative samples \cite{solon_what_2015, couper_new_2017, valliant_practical_2018}.

At a minimum, researchers should report their sampling strategies, acknowledge sampling limitations, discuss which populations results may generalize to, and discuss implications for the validity of baseline results. %

\textbf{Employ quality controls for baseliner recruitment.} Quality control (QC) mechanisms at the recruitment stage improve data quality by selecting for baseliners who can generate high-caliber data. Using inclusion/exclusion criteria during recruitment is considered best practice in survey research \cite{stantcheva_how_2023} and ensures that baseliners meet appropriate evaluation criteria. QC can include pre-testing baseliners for task-specific knowledge or general ability (see, e.g., \citealt{nangia_what_2021}) or filtering crowdsourced workers via screening questions or platform quality scores \cite{lu_improving_2022}. Of the baselines in our review, 29\% reported using quality control measures for recruitment, with most of these using pre-testing such as qualification tests or thresholds.

Depending on the research question, baseliners' domain expertise may be particularly important because expert baseliners often provide higher-quality data than non-experts \cite{cheng_human_2024, liao_are_2021}. Expert baselines are specifically needed to enable AI evaluations that compare AI performance with the ceiling of possible human performance and that explore the possibility of ``super-human'' performance \citetext{e.g., \citealt{glazer_frontiermath_2024}}.

Considerations for QC in recruitment include the cost and feasibility of recruiting (expert) baseliners, whether evaluations items require different QC criteria or expertise \citetext{cf. \citealt{weidinger_star_2024}}, and how to establish criteria for assessing baseliners (e.g., assessing domain expertise in highly specialized evaluations). Researchers may also wish to exclude baseliners who have been previously exposed to evaluation items to prevent data contamination, analogous to AI train/test contamination.

\subsection{Baseline Execution}
\label{subsec:Framework_Execution}

Baseline execution is the stage at which the human baseline is conducted and result data is collected \cite{paskov_preliminary_2025}---e.g., through surveys or crowdwork platforms. We examine four considerations for baseline execution. %

\textbf{Employ quality controls during baseline execution.} QC at the execution stage improves data quality by filtering out unreliable baseline responses. As in the recruitment stage, QC during execution is considered best practice in survey research; mechanisms include checks for attention, consistency, response pattern, outliers, and time to completion \cite{stantcheva_how_2023, lebrun_detecting_2024}. Some research has also demonstrated that attention checks may improve the representativeness of crowdwork samples \cite{qureshi_comparing_2022}. Of the baselines in our review, only 23\% reported performing QC at the execution stage, most often using attention checks and honeypot questions.

One issue of increasing importance is the inappropriate usage of AI tools by crowdworkers, which was directly raised as a concern in one article we reviewed \cite{sprague_musr_2023}. Empirical work has suggested that more than a third of MTurk and Prolific workers have used AI to complete tasks \cite{veselovsky_artificial_2023, zhang_generative_2025, traylor_threat_2025}, which can decrease data quality \cite{lebrun_detecting_2024} and baseline validity. Unintentional usage of AI tools may also occur as AI adoption increases such as via AI-generated Google Search summaries. QC to prevent AI usage may be beneficial for crowdsourced baselines: mechanisms may include explicitly asking participants not to use LLMs \cite{veselovsky_prevalence_2023}, employing technical restrictions such as preventing copy/pasting \cite{veselovsky_prevalence_2023}, using non-standard interface elements \cite{gureckis_mechanical_2021}, or using comprehension and manipulation checks \cite{frank_experimentology_2025}. Depending on the research question, AI use may be appropriate for baseliners, such as for evaluations in domains where AI usage is expected. In these cases, researchers may carefully define protocols for AI use and evaluations can compare AI capabilities with baselines of AI-augmented human capabilities \citetext{e.g., \citealt{wijk_re-bench_2024}}. %

\textbf{Control for method effects and use identical tasks.} %
Method effects are variations in item response attributable to data collection methods rather than to differences in underlying response distributions (e.g., due to instructions, option order, or mode of data collection). Method effects can reduce the internal validity of evaluations \cite{davidov_measurement_2014}. Evaluators should control for method effects wherever possible, and AI and human results should use the same tasks (i.e., identical instructions, context, etc.). Our review revealed significant discrepancies in data collection methods between humans and AI models. Of the baselines in our review, 88\% displayed UI differences between human baselining and AI evaluation, 76\% displayed differences in instructions or prompts, and 10\% displayed differences in tool access.\footnote{Our focus was on method effects between human and AI responses, but method effects can also occur between human baseliners (e.g., if baselines are collected from multiple platforms).} %

Method effects are well-documented in the social sciences, particularly in psychology and in survey methodology. Empirical research has found effects in humans due to the mode of survey administration \cite{vannieuwenhuyze_method_2010, shin_survey_2012}, question order \cite{engel_improving_2014}, fatigue from survey length \cite{stantcheva_how_2023}, example responses provided \cite{eckman_position_2025, lu_fantastically_2022}, interface design \cite{sanchez_effects_1992}, and question wording \cite{wu_confusing_2017, dafoe_information_2018}. AI systems are also subject to method effects such as prompt sensitivity and other biases \cite{anagnostidis_how_2024, ye_justice_2024}.

Measurement theory offers some guidance for addressing method effects in humans. For instance, randomization of non-critical methodological details can reduce some effects (e.g., reducing order effects by randomizing question order). Fatigue can also be addressed by shortening survey length, encouraging breaks or enforcing time limits, and implementing attention checks. %

Some method effects in AI evaluation, however, are currently inevitable due to differences between human and AI cognition \citetext{cf. \citealt{mccoy_embers_2024}}; evaluators should discuss these limitations where they could significantly affect results. For instance, many AI evaluations restart the context window for each run, but it may be unrealistic to demand that baseliners are only administered one item per sitting; only 25\% of reviewed baselines reported instrument length, of which most reported instruments were longer than one item. Similarly, although both AI systems and humans are known to be sensitive to item wording, they are sensitive in different ways \cite{tjuatja_llms_2024}, suggesting that even using the same data collection artifacts for humans and for AI systems may not prevent all method effects. %

Without clear evidence, we suggest for now that evaluators default to using identical setups for AI and human evaluation, including provision of identical instructions, examples, context, role information, and other supplementary materials or details that could affect performance (e.g., documentation, images).\footnote{These details may be particularly important to monitor when using validation data as a baseline \citetext{e.g., \textsc{GPQA} \cite{rein_gpqa_2024}}.} Researchers should also clearly document evaluation methodologies and differences in measurement instruments between AI and human results. Overall, significant additional research is needed to understand how method effects differ between humans and AI systems (and between AI systems), as well as how AI evaluations can adjust measurement instruments for these differences so as not to unfairly advantage humans or AI models in the evaluation process \cite{cowley_framework_2022, tedeschi_whats_2023}. %

\textbf{Control for level of effort.} Both humans' and AI systems' level of effort in responding to items can affect evaluation results. For AI systems, ``effort'' could be proxied by inference cost, task time limits, sampling or elicitation strategy, and other factors; analogously, baseliner effort can be affected by training, compensation, task time limits, and other factors \cite{tedeschi_whats_2023, kapoor_ai_2024}. Training could include tutorials, response guides, or example items; compensation structures can also affect baseline data quality \cite{grosse-holz_early_2024} and can vary by, e.g., payment by hour vs. per task or by performance bonuses. Of the baselines in our review, 23\% provided training to baseliners, and 41\% reported paying baseliners, with 8\% providing performance bonuses.%

Accounting for level of effort also raises design questions about the choice of the experimental unit of interest, which affects evaluations' external validity \cite{jackson_principles_2013}. Most AI evaluations take humans or AI systems as the experimental unit, but some comparisons may necessitate more granularity. For instance, \citet{wijk_re-bench_2024} compares performance after two human labor-hours vs. two AI labor-hours. Properly scoping experimental units could make evaluations more valuable for understanding AI's broader societal effects, e.g., by enabling comparisons of labor efficiency. %

\textbf{Collect qualitative data from baseliners.} Qualitative data from baseliners---including but not limited to explanations of why baseliners chose particular responses---may help interpret differences in human and AI performance, explain performance gaps, and surface validity issues. Collecting explanations is generally a best practice in survey research, as it can help surface new insights \cite{lu_improving_2022}; explanations may also be used for quality control, validation, and understanding the thought processes behind item responses \cite{lu_improving_2022, tedeschi_whats_2023}, which may lead to improvements in questions or instrumentation. Only 10\% of the baselines we reviewed collected explanations from baseliners, though this finding is unsurprising since collecting explanations may increase the cost of human baselines, and not all baselines need explanations.

\subsection{Baseline Analysis}
\label{subsec:Framework_Analysis}

Baseline analysis is the stage after data collection at which human baseline data is inspected and compared to AI results. We examine two considerations at the analysis stage. %

\textbf{Quantify uncertainty in human vs. AI performance differences.} Reporting measurements of uncertainty, such as result distributions or statistical tests, is necessary to rigorously assess whether measurements of performance truly reflect underlying performance distributions, as well as to interpreting evaluation results \cite{agarwal_deep_2022, steinbach_machine_2022, ying_benchmarking_2025}. The ML community has historically recognized these norms \citetext{e.g., \citealt{dietterich_approximate_1998, bouckaert_evaluating_2004}}, but many recent evaluations of large AI models have not met standards of statistical rigor \cite{biderman_pitfalls_2020, agarwal_deep_2022, welty_metrology_2019, paskov_gpai_2024, marie_scientific_2021}. Similarly, our review finds that only 37\% of evaluations provided interval or distribution estimates, and only 8\% performed statistical tests of any type.

Lack of statistical testing is sometimes understandable given small sample sizes and other limitations \citetext{cf. \citealt{bouthillier_accounting_2021}}.\footnote{Addressing the small sample size challenge is an ongoing area of research \cite{xiao_confidence_2025, luettgau_hibayes_2025}. See generally, \citealt{neuhauser_statistical_2024, hoyle_statistical_1999, schoot_small_2020}.} Reporting interval estimates, however, has become increasingly accessible with increased guidance \citetext{e.g., \citealt{miller_adding_2024, bowyer_position_2025}} and support in major evaluation frameworks \citetext{e.g., \citealt{uk_aisi_scorers_2025}}. Finally, in line with recent commentary in statistics, researchers should consider reporting results of statistical tests (\textit{p}-values) as one component of evidence used to judge the evaluation results, rather than as screens for statistical significance \cite{mcshane_abandon_2019, gelman_difference_2006}.

\textbf{Use consistent evaluation metrics, scoring methods, and rubrics across human and AI evaluation.} Often, comparisons between AI and human baseline results are fair only when the metrics for comparison are equivalent across samples. For instance, human baseline metrics are sometimes calculated inconsistently across items, complicating baseline interpretation \cite{tedeschi_whats_2023}; most commonly, researchers used majority vote for human but not for AI samples. Although these comparisons are not always inappropriate, researchers should consider adding clarifying language when reporting results, e.g., ``AI evaluation metrics fell below majority-vote human performance'' or ``model results on each item exceeded the maximum performance across ten human baseliners.''

\subsection{Baseline Documentation}
\label{subsec:Framework_Documentation}

Baseline documentation is the provision of evaluation tasks, datasets, metrics, and experimental materials and resources to relevant audiences \cite{reuel_betterbench_2024}. We examine two considerations for baseline documentation. %

\textbf{Report key details about baselining methodology and baseliners.} Documentation includes reporting information about baseliners, baselining procedures, and baseline paradata. Documenting methodology in particular is crucial to enable reproducibility/replicability and external assessments of baseline results. These details can significantly affect how results are contextualized, interpreted, and operationalized---especially with respect to their validity---and reporting can build collective confidence in published results \cite{liao_are_2021, biderman_lessons_2024}. 

Absent compelling reasons for confidentiality, researchers should document most of the items in our checklist that are related to baseline(r) design, recruitment, execution, and analysis (Appendix \ref{sec:Appendix_Checklist}; see also \citealt{mckee_human_2024}). Researchers should also consider reporting baseliner demographics, paradata, and other study information. Baseline demographics can enable assessments of baseliner sample representativeness and reliability; paradata such as items' time to completion can offer insights into latent variables like cognitive effort \cite{cai_item_2016, west_paradata_2011} and into data quality, which is often correlated with response times \cite{traylor_threat_2025}. Of the baselines in our review, all failed to report at least some items on our checklist, only 23\% provided detailed baseliner demographics, and only 21\% included paradata such as response times.

\textbf{Adopt best practices for open science and reproducibility/replicability.} Releasing human baseline data, experimental materials (e.g., forms, custom UIs), and analysis code in accessible repositories (e.g., GitHub, OSF) can facilitate research validation and reproduction/replication \cite{semmelrock_reproducibility_2024, stodden_best_2014}; annotator-level data is also important to gain a fuller picture of baseliner performance \cite{prabhakaran_releasing_2021}. In addition, these open science practices facilitate reuse of human baseline data in subsequent evaluations by other researchers, which in turn fosters more efficient use of resources within the ML community. Concerns around open science and replicability are not new in ML \cite{kapoor_reforms_2024, pineau_improving_2021}, and our review found that most baselines (78\%) did not publicly release human baseline responses, experimental materials (56\%), and code for analyzing human baselines (59\%).

\section{Discussion}
\label{sec:Discussion}

In this section, we discuss three additional considerations for human baselines and address the limitations of our study. %

\textbf{First, human baselines are not appropriate for all AI evaluations}. Most prominently, human baselines are not meaningful for evaluations of AI tasks without human equivalents \cite{barnett_declare_2024, laine_me_2024}. Examples include AI control evaluations, which measure an AI system's ability to monitor a more advanced AI system \cite{greenblatt_ai_2024}, and autonomous self-replication evaluations, which measure an AI agent's ability to create copies of itself \cite{pan_frontier_2024}. In contrast, human baselines can be valuable for evaluations that measure AI performance in domains with human equivalents, including but not limited to many question-answer benchmarks and task-based agent evaluations (e.g. \citealt{wijk_re-bench_2024}).

\textbf{Second, human baselines may also be constructed from secondary sources.} Our position paper focuses on primary data collection methods in human baselining, but human performance metrics can also be derived from observational/real-world data or pre-existing datasets. For instance, the Massive Multitask Language Understanding dataset uses the 95\textsuperscript{th} percentile of human standardized test scores as a point of comparison with AI results \cite{MMLU}; other studies \citetext{e.g., \citealt{hua_game-theoretic_2024}} use human subjects data from previous work \cite{lewis_deal_2017}. Re-use of human baselines highlights the need for transparency and documentation of baselining methods: authors should assume their datasets may be re-used by other researchers, who require significant methodological detail to design effective evaluations and draw meaningful conclusions from results. Secondary data is also subject to many other limitations (see Section \ref{sec:Alternative_Views}). %

\textbf{Third, human baselines can vary over time and as technology advances.} Human capabilities are known to change over time \cite{trahan_flynn_2014}, and the half-life of AI-augmented human baselines may be particularly short due to the rate of progress in AI. These trends suggest that human baselines should be interpreted as measurements at specific points in time, and researchers should tread carefully when making comparisons to older human baselines. In this vein, the ML community can consider implementing regularly updated ``living'' baselines, analogous to how public opinion polls are regularly repeated to track variation over time. Open science practices would enhance replicability and resource efficiency for such living baselines. %

\textbf{Finally, we acknowledge several limitations to our work.} Our methodology has  followed best practices for systematic literature reviews, but our meta-review sample was collected purposively and could be biased as a result (see Appendix \ref{subsec:Appendix_Meta-Review}). Our scope is limited to methodological considerations specific to human baselines, so we do not discuss in depth many important aspects of AI evaluation methodology such as construct validity \cite{strauss_construct_2009}. We also restricted our scope to foundation model evaluations; although we believe much of our framework is applicable to the broader research community, human baselines for evaluating other AI models may raise different methodological questions. Finally, future research can examine applications of measurement theory to human evaluation and human-AI interaction studies, which are not explored in this position paper (e.g. human uplift, LLM-as-a-judge).

\section{Alternative Views}
\label{sec:Alternative_Views}

We discuss four alternative views to our position below.

\textbf{Alternative View 1: Implementing all these recommendations is too expensive to be realistic.} We believe that researchers have a responsibility to ensure experimental rigor and reasonable interpretation of results; however, we also acknowledge that collecting high-quality baseline data can be prohibitively costly. Our hope is not that \textit{all} baselines will be maximally rigorous but rather that 1) \textit{all} baselines should be transparent even if not maximally rigorous; and 2) \textit{some} baselines should be both transparent and maximally rigorous. Our framework is intended to help researchers understand the impact of methodological design choices, allowing researchers to judge whether the rigor provided by particular design choices is justified by the marginal cost and by the evaluation's intended use case. Where researchers decline to opt for more rigorous methods, reporting study details is nevertheless important for transparency and can enable external assessments of published baselines. By narrowing interpretations of less rigorous baseline results and discussing limitations, researchers can also prevent readers from misunderstanding or over-hyping results. %

Moreover, we believe that many low-cost improvements can be made to existing human baselining methods, even though these improvements may not suffice for maximal rigor: e.g., using consistent test sets, satisfying ethics requirements, specifying a population of interest, using recruitment/execution QC such as excluding authors/baseliners previously exposed to evaluation items, reporting uncertainty or statistical tests, and documenting results. We also note that cost considerations are not unique to ML and have been widely acknowledged in, e.g., survey methodology \cite{leeuw_mix_2005}; ML can learn from methods in other fields such as phased clinical trials that were developed in part to account for cost. ML researchers can also collaborate with social scientists to reduce the administrative burden of gaining new expertise in measurement theory.

Finally, some baselines will need to be both transparent and maximally rigorous for use in risk management and AI governance. We believe that the value of more rigorous and transparent human baselines is sufficiently high that funders and the ML community should establish more stringent norms for scientific rigor in AI evaluations. The community should also encourage and accept individual baseline results such as \citet{nangia_human_2019} or \citet{legris_h-arc_2024} as substantial and stand-alone technical contributions.

\textbf{Alternative View 2: Human baselines will soon become unnecessary or insufficient for many evaluations as AI systems surpass expert human performance \cite{goldstein_ppou_2024}.} Human baselines---in addition to other AI baselines---may be useful even if systems surpass expert human performance. For instance, they can determine the \textit{magnitude} of human vs. AI performance differences, which is important for modeling economic impacts and for making business or policy decisions \cite{eloundou_gpts_2023}. They can also help researchers understand how cognition and behavioral tendencies differ between humans and AI systems. At the very least, a human baseline could serve as a floor for expected performance from foundation models, similar to random baselines currently. Moreover, we note that existing methods using AI to simulate human participants are subject to substantial limitations \cite{wang_large_2025, liu_large_2025, anthis_llm_2025}, so current AI systems are unlikely to be able to simulate human baseliners validly and reliably in many contexts.

\textbf{Alternative View 3: Existing human baselines or real-world data may be enough to measure progress, even if only approximately.} Some existing baselines may meaningfully measure performance, but many are insufficiently rigorous to draw conclusions about the pace of AI progress \cite{tedeschi_whats_2023, cowley_framework_2022}. Moreover, stakeholders may demand additional rigor for evaluations used in, e.g., risk assessments or safety cases \cite{goemans_safety_2024}. Secondary data like standardized tests can be useful points of comparison by providing score distributions from large samples, but it may not always exist for desired use cases. Secondary data is also less well-validated for evaluating models: data contamination concerns are common \cite{yao_data_2024}, and models can perform strangely on assessments designed for humans \citetext{e.g., \citealt{lei_gaokao-eval_2024}}.

\textbf{Alternative View 4: This framework and checklist may not be appropriate in all cases due to differing needs in human baselines.} We agree that different evaluations and contexts require different methods. Our intention is to provide a starting point for designing and assessing baselines, not necessarily a one-size-fits-all solution. Furthermore, we believe that \textit{some} standardization---common in other fields \cite{winters_clinical_2009}---is useful for transparency, replicability, and interpretability of results \citetext{see \citealt{kapoor_reforms_2024}}.

\section{Conclusion}
\label{sec:Conclusion}

In this position paper, we argue that human baselines in foundation model evaluations should be more rigorous and transparent. Systematically reviewing 115 published evaluations, we find that many baselines lack methodological rigor across the gamut of the baselining process, from design (e.g., different human vs. AI test sets) to documentation (e.g., lack of study details). We provide recommendations for human baselining based on measurement theory to foster validity and reliability, enable meaningful comparisons of human vs. AI performance, and promote research transparency. We hope that our work can guide researchers in improving baselining methods and evaluating AI systems. 

\section*{Acknowledgements}
We are extremely grateful for comments and feedback on drafts of this paper from (in randomized order): Ella Guest, three anonymous reviewers from the ICLR Workshop on Building Trust in LLMs and LLM Applications, Laura Weidinger, Michael Chen, Megan Kinniment, Fred Heiding, four anonymous ICML reviewers, Marius Hobbhahn, Christopher Summerfield, Ryan Ritterson, Lujain Ibrahim, Jacy Reese Anthis, Jacob Haimes, and Sarah Gebauer. The highly detailed comments from each of these readers have substantially improved the quality---and, we hope, the impact---of this work, and we are thankful for the time that they took to engage with our research. In addition, Anka Reuel acknowledges support from the Stanford Interdisciplinary Graduate Fellowship.

\section*{Impact Statement}

This paper presents work whose goal is to advance the field of machine learning, specifically with regards to improving the quality of methods used to create and analyze human baselines in AI evaluation. We hope that by discussing methodological considerations in human baselining---and by highlighting shortcomings in existing baselining methods---our work will lead to rigorous AI evaluations that can be useful to not just the research community but also to users of AI systems and policymakers. We discuss broader implications of human baselines in Section \ref{sec:Intro}, and we do not anticipate any particular negative impacts associated with our work.

\clearpage

\bibliography{references}
\bibliographystyle{icml2025}

\clearpage

\appendix
\onecolumn %

\clearpage

\section{Appendix: Full Results from Systematic Review} \label{sec:Appendix_Statistics}

This appendix contains summary statistics for select items from the checklist in Appendix \ref{sec:Appendix_Checklist}. 

Please note the following limitations and caveats to the statistics below. First, summary statistics are not presented for all items; all original data and additional statistics are available on Github: \url{https://github.com/kevinlwei/human-baselines}. Second, some results in this section contain imputed data; the imputation scheme is detailed in Appendix \ref{sec:Appendix_Checklist}, and individual items are marked with ``default'' responses below. Finally, additional limitations specific to each section are also noted. 

All tables below present statistics for the full set of human baselines we reviewed ($n = 115$), as well as the same statistics for a subset of human baselines from evaluations that are ($n = 7$) frequently used in industry model cards. This subset consists of: \textsc{MMMU} \cite{yue_mmmu_2024}, \textsc{GPQA} \cite{rein_gpqa_2024}, \textsc{MATH} \cite{MATH}, \textsc{DROP} \cite{dua_drop_2019}, \textsc{ARC} \cite{legris_h-arc_2024}, \textsc{ConceptARC} \cite{moskvichev_conceptarc_2023}, and \textsc{EgoSchema} \cite{mangalam_egoschema_2023}.

\setcounter{subsection}{-1}
\subsection{Paper Information}

\begin{figure}[!htbp]
    \centering
    \includegraphics[width=0.9\textwidth]{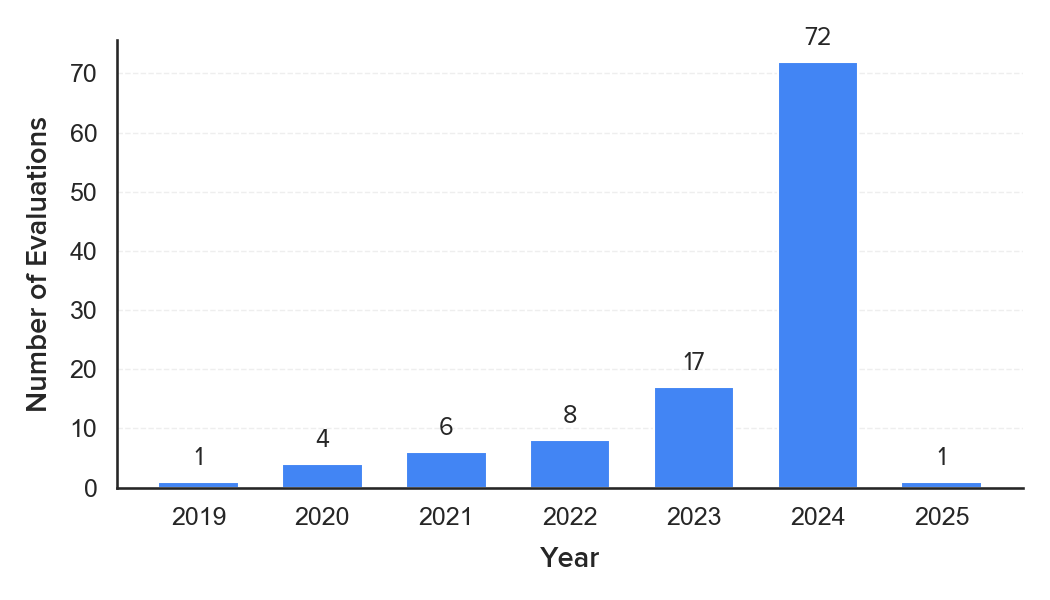}
    \caption{Frequency of years in which reviewed evaluations were published.}
    \label{fig:year-frequency}
\end{figure}

\begin{figure}[!htbp]
    \centering
    \includegraphics[width=0.9\textwidth]{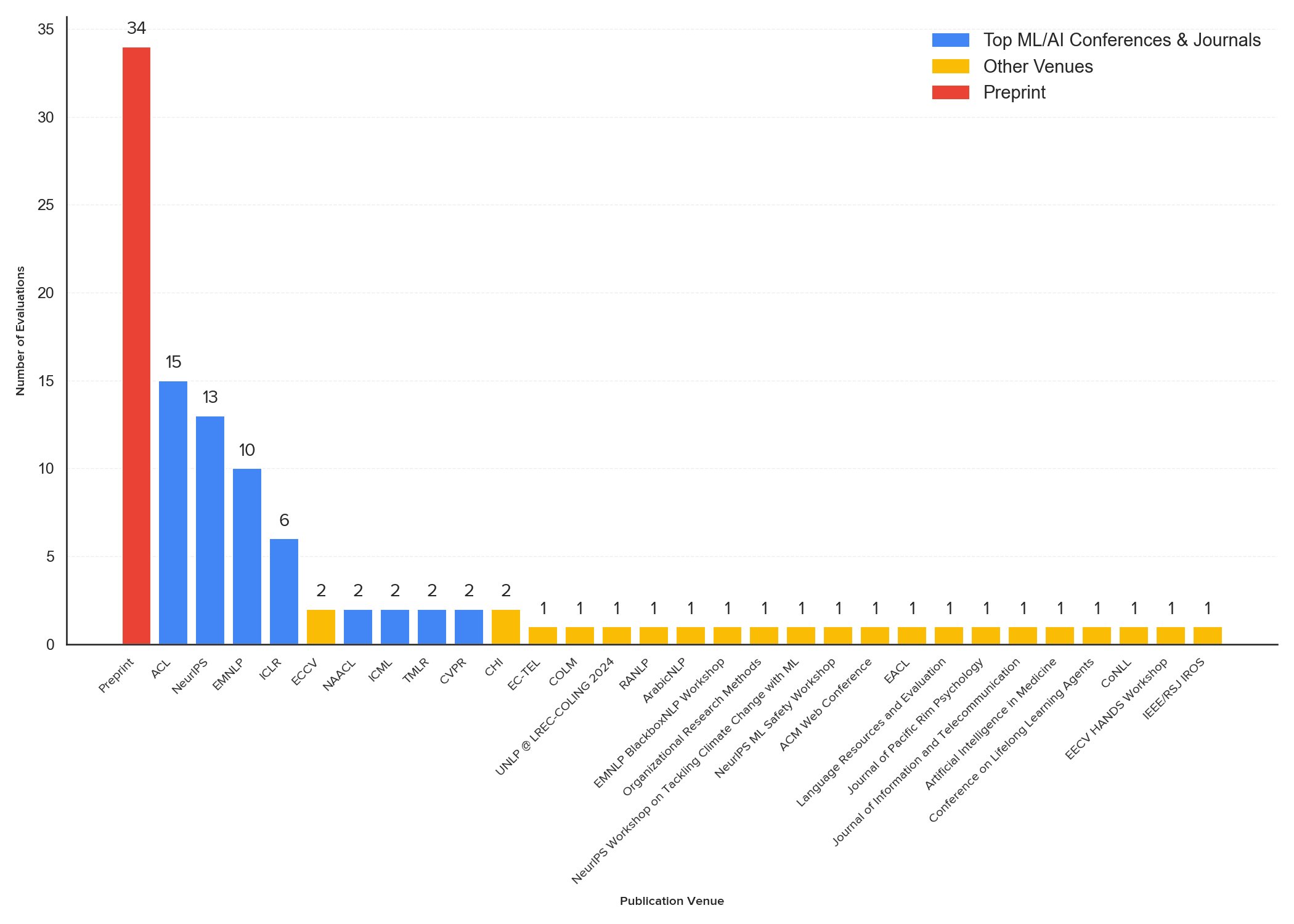}
    \caption{Frequency of publication venues of reviewed evaluations, in descending order. ``Top ML/AI conferences \& journals'' are: ICML, NeurIPS, ICLR, UAI, AISTATS, COLT, ALT, JMLR, TMLR, CVPR, ICCV, ACL, NAACL, EMNLP, and SIMODS.}
    \label{fig:venue-frequency}
\end{figure}

\begin{figure}[!htbp]
    \centering
    \includegraphics[width=0.9\textwidth]{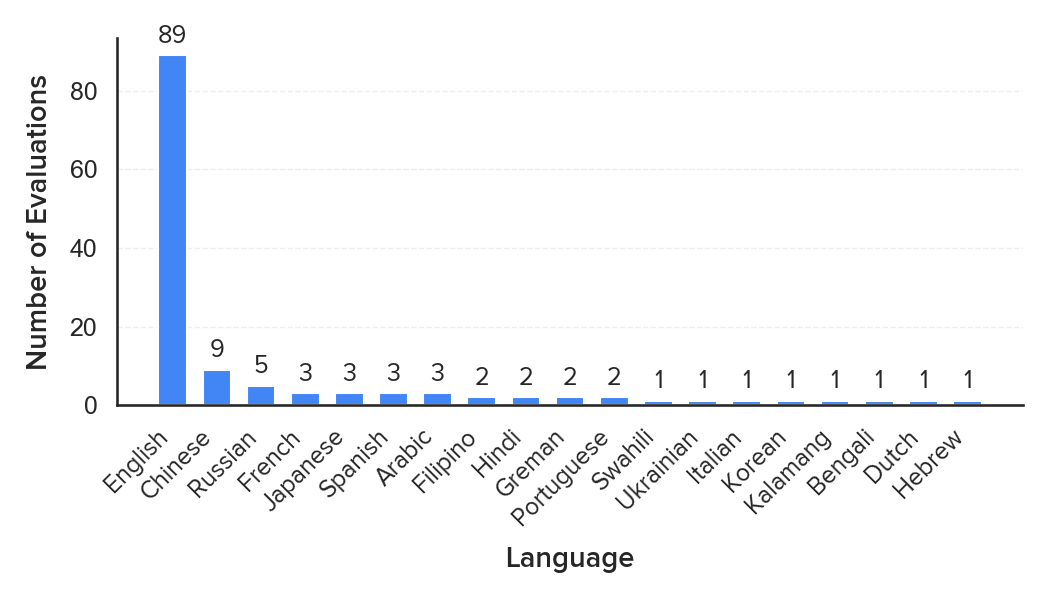}
    \caption{Frequency of languages in which reviewed evaluations' items were written, in descending order. Note that individual items may contain items in multiple languages.}
    \label{fig:language-frequency}
\end{figure}

Figures \ref{fig:year-frequency}--\ref{fig:language-frequency} contain frequency charts for the years, publication venues, and languages of reviewed evaluations. Note that the list of ``top ML/AI conferences \& journals'' in Figure \ref{fig:venue-frequency} is taken from the ICML Reviewer FAQs (i.e., the list of venues for which ICML reviewers are expected to have published in); \footnote{\url{https://perma.cc/D6TK-YQFJ}.} we understand that conference and journal ``rankings'' are inherently subjective and do not take a position on which publication venues are objectively ``better'' or ``worse.'' Figure \ref{fig:venue-frequency} also contains workshops and non-archival venues.

\clearpage

\subsection{Baseline Design \& Implementation}

NB: the ``test set equivalence'' row in Table \ref{tab:stats_di_imputed} is not a separate item in our checklist but rather is imputed from the number of evaluation items in the AI test set and the human test set, as well as an item (not reported below but available on our Github\footnote{\url{https://github.com/kevinlwei/human-baselines}.}) about whether comparisons between human and AI performance are made on the same test set.

\begin{table}[!htbp]
    \centering
    \small
    \begin{tabularx}{\textwidth}{ p{0.40\textwidth} YY YY }
        \toprule
        \multirow{2}{*}{\textbf{Question}} & \multicolumn{2}{c}{\textbf{All Baselines} ($n = 115$)} & \multicolumn{2}{c}{\textbf{Model Card Baselines} ($n = 7$)} \\
         & Yes & No
         & Yes & No
         \\
         
        \midrule

        \textbf{Test Set Equivalence}: Were human and AI test sets identical? (Default: No) & 
            \makecell[t]{59.13\% \\ 68} & 
            \makecell[t]{40.87\% \\ 47} &
            \makecell[t]{57.14\% \\ 4} &
            \makecell[t]{42.86\% \\ 3} 
            \\
            
        1.5 \textbf{Explicit Human/AI Adjustment}: Does the eval/baseline instructions and items account for both humans and AI models completing the evals items (questions/tasks)? (Default: No) & 
            \makecell[t]{16.52\% \\ 19} & 
            \makecell[t]{83.48 \\ 96} &
            \makecell[t]{28.57\% \\ 2} &
            \makecell[t]{71.43\% \\ 5} 
            \\

        1.8 \textbf{Power Analysis}: Did the authors conduct power analysis in order to determine baseline size? (Default: No) & 
            \makecell[t]{1.74\% \\ 2} & 
            \makecell[t]{98.26\% \\ 113} &
            \makecell[t]{0.00\% \\ 0} &
            \makecell[t]{100.00\% \\ 7} 
            \\

        1.10 \textbf{Pre-Registration}: Was the baseline/eval design pre-registered? (Default: No) & 
            \makecell[t]{1.74\% \\ 2} & 
            \makecell[t]{98.26\% \\ 113} &
            \makecell[t]{0.00\% \\ 0} &
            \makecell[t]{100.00\% \\ 7} 
            \\

        \bottomrule
    \end{tabularx}
    \caption{Summary statistics for baseline design \& implementation items (with imputation)} \label{tab:stats_di_imputed}
\end{table}

\begin{table}[!htbp]
    \centering
    \small
    \begin{tabularx}{\textwidth}{ p{0.4\textwidth} YYY YYY }
        \toprule
        
        \multirow{2}{*}{\textbf{Question}} & \multicolumn{3}{c}{\textbf{All Baselines} ($n = 115$)} & \multicolumn{3}{c}{\textbf{Model Card Baselines} ($n = 7$)} \\
         & Yes & No & Unknown
         & Yes & No & Unknown
         \\
         
        \midrule

        1.6 \textbf{Iterative Design}: Was the experimental setup of the baseline iteratively designed with participatory methods? & 
            \makecell[t]{34.78\% \\ 40} &
            \makecell[t]{12.17\% \\ 14} &
            \makecell[t]{63.04\% \\ 61} &
            \makecell[t]{42.86\% \\ 3} &
            \makecell[t]{28.57\% \\ 2} &
            \makecell[t]{28.57\% \\ 2} 
            \\
        
        1.7 \textbf{Amount of Effort}: Does the baseline control for the amount of effort by human baseliners and AIs & 
            \makecell[t]{13.91\% \\ 16} & 
            \makecell[t]{35.65\% \\ 41} &
            \makecell[t]{50.43\% \\ 58} & 
            \makecell[t]{28.57\% \\ 2} &
            \makecell[t]{57.14\% \\ 4} &
            \makecell[t]{14.29 \\ 1} 
            \\

        1.9 \textbf{Ethics Review}: Was the study approved or exempted by an IRB, or did it undergo other ethics review?  & 
            \makecell[t]{13.91\% \\ 16} & 
            \makecell[t]{2.61\% \\ 3} &
            \makecell[t]{83.48\% \\ 96} & 
            \makecell[t]{0.00\% \\ 0} &
            \makecell[t]{14.29\% \\ 1} &
            \makecell[t]{85.71\% \\ 6} 
            \\
            
        \bottomrule
    \end{tabularx}
    \caption{Summary statistics for baseline design \& implementation items (no)}
\end{table}

\clearpage

\subsection{Baseliner Recruitment}

NB: note that our statistics for Q2.1 are likely significant over-estimates, as we erred on the side of more generous annotations. Most papers did not explicitly specify populations of interest; papers that gestured at baseline demographics (e.g., ''our baseliners were 18--25 years old'') were assumed to have specified a population of interest defined by those demographics (e.g., the population in the previous example would be all adults 18--25 years of age).

\begin{table}[!htbp]
    \centering
    \small
    \begin{tabularx}{\textwidth}{ p{0.4\textwidth} YY YY }
        \toprule
        \multirow{2}{*}{\textbf{Question}} & \multicolumn{2}{c}{\textbf{All Baselines} ($n = 115$)} & \multicolumn{2}{c}{\textbf{Model Card Baselines} ($n = 7$)} \\
         & Yes & No
         & Yes & No
         \\
         
        \midrule
        2.1 \textbf{Population of Interest Identification}: Does the reporting identify human populations for which these results may be valid, i.e., a human population of interest? (Default: No) & 
            \makecell[t]{42.61\% \\ 49} & 
            \makecell[t]{57.39\% \\ 66} &
            \makecell[t]{57.14\% \\ 4} &
            \makecell[t]{42.86\% \\ 3} 
            \\

        2.3 \textbf{Quality Control in Recruitment}: Were human baseliners pre-qualified or excluded during the recruitment process for any reason? (Default: Yes) & 
            \makecell[t]{28.70\% \\ 33} & 
            \makecell[t]{71.30\% \\ 82} &
            \makecell[t]{28.57\% \\ 2} &
            \makecell[t]{71.43\% \\ 5} 
            \\

        2.4 \textbf{Author Baseliners}: Did the authors or members of the research team also serve as human baseliners? (Default: No) & 
            \makecell[t]{9.33\% \\ 14} & 
            \makecell[t]{91.67\% \\ 101} &
            \makecell[t]{28.57\% \\ 2} &
            \makecell[t]{71.43\% \\ 5} 
            \\

        2.5 \textbf{Baseliner Train/Test Contamination}: Did the recruitment process exclude baseliners who had been exposed to the eval questions previously? (Default: No) & 
            \makecell[t]{7.83\% \\ 9} & 
            \makecell[t]{92.17\% \\ 106} &
            \makecell[t]{14.29\% \\ 1} &
            \makecell[t]{85.71\% \\ 6} 
            \\

        2.6 \textbf{Baseliner Training}: Did the human baseliners receive training for the baseline? (Default: No) & 
            \makecell[t]{22.61\% \\ 26} & 
            \makecell[t]{77.39\% \\ 89} &
            \makecell[t]{42.86\% \\ 3} &
            \makecell[t]{57.14\% \\ 4} 
            \\

        \bottomrule
    \end{tabularx}
    \caption{Summary statistics for baseliner recruitment items (with imputation)}
\end{table}

\begin{table}[!htbp]
    \centering
    \small
    \begin{tabularx}{\textwidth}{ p{0.35\textwidth} YYY YYY }
        \toprule
        
        \multirow{2}{*}{\textbf{Question}} & \multicolumn{3}{c}{\textbf{All Baselines} ($n = 115$)} & \multicolumn{3}{c}{\textbf{Model Card Baselines} ($n = 7$)} \\
         & Convenience & Crowdsource & Unknown
         & Convenience & Crowdsource & Unknown
         \\
         
        \midrule
        2.2 \textbf{Baseliner Sampling Strategy}: How were the human baseliners recruited? & 
            \makecell[t]{31.30\% \\ 36} & 
            \makecell[t]{32.17\% \\ 37} &
            \makecell[t]{36.52\% \\ 42} & 
            \makecell[t]{28.57\% \\ 2} &
            \makecell[t]{42.86\% \\ 3} &
            \makecell[t]{28.57\% \\ 2} 
            \\
            
        \bottomrule
    \end{tabularx}
    \caption{Summary statistics for Q2.2 Baseliner Sampling Strategy (no imputation)}
\end{table}

\begin{table}[!htbp]
    \centering
    \small
    \begin{tabularx}{\textwidth}{ p{0.4\textwidth} YYY YYY }
        \toprule
        
        \multirow{2}{*}{\textbf{Question}} & \multicolumn{3}{c}{\textbf{All Baselines} ($n = 115$)} & \multicolumn{3}{c}{\textbf{Model Card Baselines} ($n = 7$)} \\
         & Yes & No & Unknown
         & Yes & No & Unknown
         \\
         
        \midrule
        2.7 \textbf{Baseliner Testing Compensation}: Were the human baseliners compensated for completing the baseline? (Default: No) & 
            \makecell[t]{41.74\% \\ 48} & 
            \makecell[t]{11.03\% \\ 13} &
            \makecell[t]{46.96\% \\ 54} & 
            \makecell[t]{57.14\% \\ 4} &
            \makecell[t]{0.00\% \\ 0} &
            \makecell[t]{42.86\% \\ 3} 
            \\
            
        \bottomrule
    \end{tabularx}
    \caption{Summary statistics for Q2.7 Baseliner Testing Compensation (no imputation)}
\end{table}

\clearpage

\subsection{Baseline Execution}

\begin{table}[!htbp]
    \centering
    \small
    \begin{tabularx}{\textwidth}{ p{0.40\textwidth} YY YY }
        \toprule
        \multirow{2}{*}{\textbf{Question}} & \multicolumn{2}{c}{\textbf{All Baselines} ($n = 115$)} & \multicolumn{2}{c}{\textbf{Model Card Baselines} ($n = 7$)} \\
         & Yes & No
         & Yes & No
         \\
         
        \midrule
        3.2 \textbf{Quality Control in Execution}: Were quality checks implemented or data cleaned/excluded during the data collection process (i.e., after baseliners were recruited)? (Default: No) & 
            \makecell[t]{23.48\% \\ 27} & 
            \makecell[t]{76.52\% \\ 88} &
            \makecell[t]{28.57\% \\ 2} &
            \makecell[t]{71.43\% \\ 5} 
            \\

        3.3 \textbf{UI Equivalence}: Did the human baseliners and AIs have access to the same UI for each item? (Default: No) & 
            \makecell[t]{12.17\% \\ 14} & 
            \makecell[t]{87.83\% \\ 101} &
            \makecell[t]{14.29\% \\ 1} &
            \makecell[t]{85.71\% \\ 6} 
            \\

        3.4 \textbf{Instruction Equivalence}: Did the human baseliners and AIs have access to the same instructions/prompt/question for each item? (Default: No) & 
            \makecell[t]{24.35\% \\ 28} & 
            \makecell[t]{75.65\% \\ 87} &
            \makecell[t]{14.29\% \\ 1} &
            \makecell[t]{85.71\% \\ 6} 
            \\

        3.5 \textbf{Tool Access Equivalence}: Did the human baseliners and AIs have access to the same (technical) tools for each item? (Default: Yes) & 
            \makecell[t]{89.57\% \\ 103} & 
            \makecell[t]{10.43\% \\ 12} &
            \makecell[t]{71.43\% \\ 5} &
            \makecell[t]{28.57\% \\ 2} 
            \\

        3.6 \textbf{Explanations}: Did the eval/baseline collect explanations from the human baseliners, after the evaluation was conducted? (Default: No) & 
            \makecell[t]{11.30\% \\ 13} & 
            \makecell[t]{97.39\% \\ 112} &
            \makecell[t]{42.86\% \\ 3} &
            \makecell[t]{57.14\% \\ 4} 
            \\

        \bottomrule
    \end{tabularx}
    \caption{Summary statistics for baseline execution items (with imputation)}
\end{table}

\clearpage

\subsection{Baseline Analysis}

\begin{table}[!htbp]
    \centering
    \small
    \begin{tabularx}{\textwidth}{ p{0.40\textwidth} YY YY }
        \toprule
        \multirow{2}{*}{\textbf{Question}} & \multicolumn{2}{c}{\textbf{All Baselines} ($n = 115$)} & \multicolumn{2}{c}{\textbf{Model Card Baselines} ($n = 7$)} \\
         & Yes & No
         & Yes & No
         \\
         
        \midrule
        4.1 \textbf{Statistical Significance}: Did the eval test for statistically significant differences between AI and human performance? (Default: No) & 
            \makecell[t]{8.70\% \\ 10} & 
            \makecell[t]{91.30\% \\ 105} &
            \makecell[t]{0.00\% \\ 0} &
            \makecell[t]{100.00\% \\ 7} 
            \\

        4.2 \textbf{Uncertainty Estimate}: Did the paper present a measure of uncertainty for the AI and human baseline results? (Default: No) & 
            \makecell[t]{33.04\% \\ 38} & 
            \makecell[t]{66.96\% \\ 77} &
            \makecell[t]{14.29\% \\ 1} &
            \makecell[t]{85.71\% \\ 7} 
            \\

        4.3 \textbf{Evaluation Metric Equivalence}: Was the same evaluation metric measured/compared for both humans and AIs? (Default: Yes) & 
            \makecell[t]{93.91\% \\ 108} & 
            \makecell[t]{6.09\% \\ 7} &
            \makecell[t]{100.00\% \\ 7} &
            \makecell[t]{0.00\% \\ 0} 
            \\

        4.4 \textbf{Evaluation Scoring Criteria Equivalence}: Was the same scoring rubric used for both AI and human results? (Default: Yes) & 
            \makecell[t]{98.26\% \\ 113} & 
            \makecell[t]{1.74\% \\ 2} &
            \makecell[t]{100.00\% \\ 7} &
            \makecell[t]{0.00\% \\ 0} 
            \\

        4.5 \textbf{Evaluation Scoring Method Equivalence}: Was the same scoring method used for both AI and human results? (Default: Yes) & 
            \makecell[t]{95.65\% \\ 110} & 
            \makecell[t]{4.35\% \\ 5} &
            \makecell[t]{100.00\% \\ 7} &
            \makecell[t]{0.00\% \\ 0} 
            \\

        \bottomrule
    \end{tabularx}
    \caption{Summary statistics for baseline analysis items (with imputation)}
\end{table}

\begin{table}[!htbp]
    \centering
    \small
    \begin{tabularx}{\textwidth}{ p{0.4\textwidth} YYY YYY }
        \toprule
        \multirow{2}{*}{\textbf{Question}} & \multicolumn{3}{c}{\textbf{All Baselines} ($n = 115$)} & \multicolumn{3}{c}{\textbf{Model Card Baselines} ($n = 7$)} \\
         & Point & Interval & Distribution & Point & Interval & Distribution \\
        \midrule
        
        4.2.1 \textbf{Estimate Type}: Is the reported baseline a point estimate, an interval estimate, or a distribution? (Default: Point Estimate) & 
        \makecell[t]{63.48\% \\ 73} & 
        \makecell[t]{32.17\% \\ 37} &
        \makecell[t]{4.35\% \\ 5} &
        \makecell[t]{71.43\% \\ 5} &
        \makecell[t]{14.29\% \\ 1} &
        \makecell[t]{14.29\% \\ 1} 
        \\
        
        \bottomrule
    \end{tabularx}
    \caption{Summary statistics for Q4.2.1 Estimate Type (with imputation)}
\end{table}

\clearpage

\subsection{Baseline Documentation}

\begin{table}[!htbp]
    \centering
    \small
    \begin{tabularx}{\textwidth}{ p{0.40\textwidth} YY YY }
        \toprule
        \multirow{2}{*}{\textbf{Question}} & \multicolumn{2}{c}{\textbf{All Baselines} ($n = 115$)} & \multicolumn{2}{c}{\textbf{Model Card Baselines} ($n = 7$)} \\
         & Yes & No
         & Yes & No
         \\
         
        \midrule
        5.1.1 \textbf{Reporting Sample Demographics}: Were demographics for human baseliners, e.g., race, gender, etc. reported? (Default: No) & 
            \makecell[t]{22.61\% \\ 26} & 
            \makecell[t]{77.39\% \\ 89} &
            \makecell[t]{28.57\% \\ 2} &
            \makecell[t]{71.43\% \\ 5} 
            \\

        5.1.2 \textbf{Reporting Baseline Instructions}: Were instructions/guidelines given to human baseliners reported? (Default: No) & 
            \makecell[t]{40.00\% \\ 46} & 
            \makecell[t]{60.00\% \\ 69} &
            \makecell[t]{42.86\% \\ 3} &
            \makecell[t]{57.14\% \\ 4} 
            \\

        5.1.3 \textbf{Reporting Time to Completion}: Was time to completion for eval items reported? (Default: No) & 
            \makecell[t]{20.87\% \\ 24} & 
            \makecell[t]{79.13\% \\ 91} &
            \makecell[t]{28.57\% \\ 2} &
            \makecell[t]{71.43\% \\ 5} 
            \\

        5.2 \textbf{Baseline Data Availability}: Is the (anonymized) human baseline data publicly available? (Default: No) & 
            \makecell[t]{21.74\% \\ 25} & 
            \makecell[t]{78.26\% \\ 90} &
            \makecell[t]{28.57\% \\ 2} &
            \makecell[t]{71.43\% \\ 5} 
            \\
        
        5.3 \textbf{Experimental Materials Availability}: Are experimental materials used to implement the eval/baseline publicly available? (Default: No) & 
            \makecell[t]{46.97\% \\ 54} & 
            \makecell[t]{55.65\% \\ 64} &
            \makecell[t]{57.14\% \\ 4} &
            \makecell[t]{42.86\% \\ 3} 
            \\

        5.4 \textbf{Analysis Code Availability}: Is the code used to analyze the eval/baseline publicly available? (Default: No) & 
            \makecell[t]{40.87\% \\ 47} & 
            \makecell[t]{59.13\% \\ 56} &
            \makecell[t]{57.14\% \\ 4} &
            \makecell[t]{42.86\% \\ 3} 
            \\
        \bottomrule
    \end{tabularx}
    \caption{Summary statistics for baseline documentation items (with imputation)}
\end{table}

\clearpage

\section{Appendix: Full Checklist}
\label{sec:Appendix_Checklist}

Our checklist is presented in full below, updated with slight modifications and reorganization from the version used in our coding process. Our hope is that this checklist can guide and inform researchers in building human baselines and in reporting baseline results. 

Note that the following changes were made during our coding process:

\begin{itemize}[itemsep=0pt, topsep=0pt]
    \item All items were open text fields unless explicitly indicated otherwise below. 
    \item For questions on a scale of ``Yes'', ``Partial'', ``No'', ``Unknown/Unreported'', or ``N/A'': 
    \begin{itemize}
        \item ``Yes'' and ``No'' options were selected only if the relevant checklist item was explicitly noted in an article's main text, supplementary material/appendices, or GitHub codebase.
        \item ``Partial'' was selected where articles did not fully satisfy the item criterion, e.g., satisfying the criterion for some but not all of the baseline items. ``Partial'' labels were ``rounded'' up to ``Yes'' labels unless otherwise specified below.
        \item ``Unknown/Unreported'': see below. 
        \item ``N/A'' was selected where the item did not apply to the baseline at hand.
    \end{itemize}
    \item For all questions, including items with open text fields: coders indicated ``Unknown/Unreported'' where items were not reported or where coders were not able to determine the response based on an article's main text, supplementary material/appendices, or GitHub codebase. 
    \begin{itemize}
        \item For select items, ``Unknown/Unreported'' labels were imputed---i.e., resolved to default values, which are indicated below in \ul{underline} and with a ``\ul{(Default)}'' label. Default responses are selected based on our understanding of common practices in AI evaluation, and we attempt to be liberal in terms of assuming rigor in the baseline where there is no consensus in the literature on common practices.
        \item For items without default responses, ``Unknown/Unreported'' labels were not adjusted.
    \end{itemize}
\end{itemize}

\setcounter{subsection}{-1}
\subsection{Paper Information}

\renewcommand{\labelenumi}{0.\arabic{enumi}}
\renewcommand{\labelenumii}{0.\arabic{enumi}.\arabic{enumii}}
\renewcommand{\labelenumiii}{0.\arabic{enumi}.\arabic{enumii}.\arabic{enumiii}}

\begin{enumerate}[leftmargin=30pt, topsep=0pt, itemsep=0pt]
    \item \textbf{Paper Title}
    
    \item \textbf{Paper Link}
    
    \item \textbf{Publication Year}
    
    \item \textbf{Publication Venue}
    
    \item \textbf{Type of Eval} 
    \newline \textit{Select all that apply}
    \begin{itemize}
        \item Knowledge
        \item Capabilities
        \item Propensity
        \item Agent
    \end{itemize}
    
    \item \textbf{Mode of Eval}
    \newline \textit{Select all that apply}
    \begin{itemize}
        \item Text
        \item Visual (photo/video)
        \item Audio
        \item Other
    \end{itemize}
    
    \item \textbf{Language of Eval} 
    \newline \textit{Select all that apply from list}
    
    \item \textbf{Evaluation Dataset Size}: What is the total number of items in the evaluation dataset? 
    
    \item \textbf{AI Test Set Size}: What is the number of items that the AI evaluation is run on? \ul{(Default same as Q0.8)}
    
    \item \textbf{AI Samples per Item}: What is the number of AI responses (``samples'' or ``runs'') that is collected for each item? \ul{(Default 1)} 
\end{enumerate}

\subsection{Baseline Design \& Implementation}

\renewcommand{\labelenumi}{1.\arabic{enumi}}
\renewcommand{\labelenumii}{1.\arabic{enumi}.\arabic{enumii}}
\renewcommand{\labelenumiii}{1.\arabic{enumi}.\arabic{enumii}.\arabic{enumiii}}

\begin{enumerate}[leftmargin=30pt, topsep=0pt, itemsep=0pt]
    \item \textbf{Number of Baseliners}: How many baseliners were there total?
    
    \item \textbf{Baseline Test Set Size}: What is the number of items that the human baseline is run on? (i.e., how many of the questions do the baseliners collectively answer?) \ul{(Default same as Q0.9)}
    \begin{enumerate}
        \item \textbf{Baseline Test Set Sampling Strategy}: If the baseline is only run on a sample of the total dataset: what is the sampling strategy behind how the items were selected? E.g., simple random sampling, stratified sampling, etc.
    \end{enumerate}
    
    \item \textbf{Baseline Samples per Item}: What was the number of human baseliner responses that is collected for each item? \ul{(Default Q1.1 $*$ Q1.4 $/$ Q1.2, or 1 if Q1.1 or Q1.4 unreported)}
    
    \item \textbf{Items per Baseliner}: What is the number of items that each baseliner responded to?
    
    \item \textbf{Explicit Human/AI Adjustment}: Does the eval/baseline instructions and items account for both humans and AI models completing the evals items (questions/tasks)? E.g., do the authors of the eval explicitly state that the eval is designed so as not to advantage either humans or AI models? 
    \newline \textit{Select one of: ``Yes'', ``Partial'', \ul{``No'' (Default)}, ``Unknown/Unreported'', or ``N/A''}
    
    \item \textbf{Iterative Design}: Was the experimental setup of the baseline iteratively designed with participatory methods? E.g., was there a pilot study, expert validation of the items, etc.? 
    \newline \textit{Select one of: ``Yes'', ``Partial'', ``No'', ``Unknown/Unreported'', or ``N/A''}
    
    \item \textbf{Amount of Effort}: Does the baseline control for the amount of effort by human baseliners and AIs? E.g., in terms of cost, time, etc.
    \newline \textit{Select one of: ``Yes'', ``Partial'', ``No'', ``Unknown/Unreported'', or ``N/A''}
    
    \item \textbf{Power Analysis}: Did the authors conduct power analysis in order to determine baseline size? 
    \newline \textit{Select one of: ``Yes'', ``Partial'', \ul{``No'' (Default)}, ``Unknown/Unreported'', or ``N/A''}
    
    \begin{enumerate}
        \item \textbf{Minimum Detectable Effect Size}: if yes, what is the minimum detectable effect size and power?
    \end{enumerate}
    
    \item \textbf{Ethics Review}: Was the study approved or exempted by an IRB, or did it undergo other ethics review? 
    \newline \textit{Select one of: ``Yes'', ``Partial'', ``No'', ``Unknown/Unreported'', or ``N/A''}
    
    \item \textbf{Pre-Registration}: Was the baseline/eval design pre-registered? I.e., a plan detailing the experimental setup that is publicly registered online before running the experiment (e.g., on OSF, COS, etc.) 
    \newline \textit{Select one of: ``Yes'', ``Partial'', \ul{``No'' (Default)}, ``Unknown/Unreported'', or ``N/A''}
\end{enumerate}

\subsection{Baseliner Recruitment}

\renewcommand{\labelenumi}{2.\arabic{enumi}}
\renewcommand{\labelenumii}{2.\arabic{enumi}.\arabic{enumii}}
\renewcommand{\labelenumiii}{2.\arabic{enumi}.\arabic{enumii}.\arabic{enumiii}}

\begin{enumerate}[leftmargin=30pt, topsep=0pt, itemsep=0pt]
    \item \textbf{Population of Interest Identification}: Does the reporting identify human populations for which these results may be valid, i.e., a human population of interest?
    \newline \textit{Select one of: ``Yes'', ``Partial'', \ul{``No'' (Default)}, ``Unknown/Unreported'', or ``N/A''}
    \begin{enumerate}
        \item \textbf{Population of Interest Identification Criteria}: Which of the following factors were used to scope the target human population of interest?
        \newline \textit{Select all that apply}
        \begin{itemize}
            \item Expertise
            \item Education
            \item Language
            \item Gender/sex
            \item Race
            \item Socioeconomic status
            \item Age
            \item Disabilities/impairments
            \item Political orientation
            \item Digital literacy (Prior experience with computers)
            \item AI literacy (Prior experience with AI tools)
            \item Baseline experience: Prior experience with AI evals/doing human baselines 
            \item Other (specify)
        \end{itemize}
    \end{enumerate}
    
    \item \textbf{Baseliner Sampling Strategy}: How were the human baseliners recruited? 
    \newline \textit{Select one of the below}
    \begin{itemize}
        \item Crowdsource
        \item Convenience sample
        \item Simple random sample
        \item Stratified random sample
        \item Other (specify)
        \item Unknown/unreported
    \end{itemize}
    
    \item \textbf{Quality Control in Recruitment}: Were human baseliners pre-qualified or excluded during the recruitment process for any reason?
    \newline \textit{Select one of: \ul{``Yes'' (Default)}, ``Partial'', ``No'', ``Unknown/Unreported'', or ``N/A''}
    \begin{enumerate}
        \item \textbf{Quality Control Criteria for Baseliners}: If yes: please describe the inclusion/exclusion criteria for human baseliners (e.g., pre-tests, expert judgements/filtering, quality scores or ratings on crowdwork platforms, number of tasks completed on crowdwork platforms). Data quality checks that occurred after baseliners were recruited should be reported in the implementation section (e.g., attention checks in a survey).
        
        \item \textbf{Recruitment Exclusion Rate}: If yes: how many baseliners were excluded from the final baseline based on these criteria?
    \end{enumerate}
        
    \item \textbf{Author Baseliners}: Did the authors or members of the research team also serve as human baseliners?
    \newline \textit{Select one of: ``Yes'', ``Partial'', \ul{``No'' (Default)}, ``Unknown/Unreported'', or ``N/A''}
    
    \item \textbf{Baseliner Train/Test Contamination}: Did the recruitment process exclude baseliners who had been exposed to the eval questions previously?
    \newline \textit{Select one of: ``Yes'', ``Partial'', \ul{``No'' (Default)}, ``Unknown/Unreported'', or ``N/A''}
    
    \item \textbf{Baseliner Training}: Did the human baseliners receive training for the baseline? Training should be distinct from the reported data, e.g., a tutorial completed before answering baseline questions
    \newline \textit{Select one of: ``Yes'', ``Partial'', \ul{``No'' (Default)}, ``Unknown/Unreported'', or ``N/A''}
    \begin{enumerate}
        \item \textbf{Baseliner Training Type}: If yes: describe the type of training received (e.g., tutorial, shown examples, etc.)
        \item \textbf{Baseliner Training Compensation}: If yes: were the baseliners compensated for the training?
        \newline \textit{Select one of: ``Yes'', ``Partial'', ``No'', ``Unknown/Unreported'', or ``N/A''}
        
        \begin{enumerate}
            \item \textbf{Baseliner Training Compensation Amount}: If yes: list the compensation per baseliner (preferably \$ / hour, otherwise total \$ amount if stated)
        \end{enumerate}
    \end{enumerate}
    
    \item \textbf{Baseliner Testing Compensation}: Were the human baseliners compensated for completing the baseline?
    \newline \textit{Select one of: ``Yes'', ``Partial'', ``No'' (Default), ``Unknown/Unreported'', or ``N/A''}
    \begin{enumerate}
        \item \textbf{Baseliner Testing Compensation Amount}: If yes: how much was compensation? (preferably \$ / hour, otherwise total \$ amount if stated)

        \item \textbf{Baseliner Testing Performance Bonus}: If yes: was a performance bonus offered to baseliners?
        \newline \textit{Select one of: \ul{``Yes'' (Default)}, ``Partial'', ``No'', ``Unknown/Unreported'', or ``N/A''}

        \begin{enumerate}
            \item \textbf{Baseliner Testing Performance Bonus Amount}: If yes: how much was the performance bonus, and how was it determined?
        \end{enumerate}
        
        \item \textbf{Baseliner Testing Compensation Structure}: If yes: were compensation rates and structures constant across baseliners? E.g., respond no if baseliners were paid differently according to expertise.
        \newline \textit{Select one of: ``Yes'', ``Partial'', ``No'', ``Unknown/Unreported'', or ``N/A''}
        \begin{enumerate}
            \item \textbf{Baseliner Testing Compensation Structure Details}: If not compensated equally: how were compensation amounts determined?
        \end{enumerate}
    \end{enumerate}
\end{enumerate}

\subsection{Baseline Execution}

\renewcommand{\labelenumi}{3.\arabic{enumi}}
\renewcommand{\labelenumii}{3.\arabic{enumi}.\arabic{enumii}}
\renewcommand{\labelenumiii}{3.\arabic{enumi}.\arabic{enumii}.\arabic{enumiii}}

\begin{enumerate}[leftmargin=30pt, topsep=0pt, itemsep=0pt]
    \item \textbf{Instrument Length}: How many items did the human baseliners complete in a single sitting/session? I.e., what is the length of the baseliner ``context window'' in units of items?
    \begin{enumerate}
        \item \textbf{Item Randomization}: If not 1: was the order of the questions randomized?
    \end{enumerate}
    
    \item \textbf{Quality Control in Execution}: Were quality checks implemented or data cleaned/excluded during the data collection process (i.e., after baseliners were recruited)? E.g., were there any exclusion criteria for baseliner responses due to data quality such as attention check questions, honeypot questions, filtering out responders who completed the eval too quickly, screen recording, etc.
    \newline \textit{Select one of: ``Yes'', ``Partial'', \ul{``No'' (Default)}, ``Unknown/Unreported'', or ``N/A''}
    \begin{enumerate}
        \item \textbf{Quality Control in Execution Criteria}: If yes: what factors were used to determine data quality or to exclude low-quality data?
        
        \item \textbf{Execution Exclusion Rate}: If yes: how many samples were excluded from the final baseline based on these criteria?
    \end{enumerate}
    
    \item \textbf{UI Equivalence}: Did the human baseliners and AIs have access to the same UI for each item?
    \newline \textit{Select one of: ``Yes'', ``Partial'', \ul{``No'' (Default)}, ``Unknown/Unreported'', or ``N/A''}
    \begin{enumerate}
        \item \textbf{GUI vs. API}: Check this box if the humans had access to a graphical UI and the AIs only had API inputs
        \newline \textit{Checkbox item} \ul{(Unchecked by default)}
        
        \item \textbf{UI Equivalence Adjustment}: If no: does the eval attempt to adjust for the differences?
        \newline \textit{Select one of: ``Yes'', ``Partial'', \ul{``No'' (Default)}, ``Unknown/Unreported'', or ``N/A''}
    \end{enumerate}
    
    \item \textbf{Instruction Equivalence}: Did the human baseliners and AIs have access to the same instructions/prompt/question for each item?
    \newline \textit{Select one of: ``Yes'', ``Partial'', \ul{``No'' (Default)}, ``Unknown/Unreported'', or ``N/A''}
    \begin{enumerate}
        \item \textbf{Instruction Equivalence Adjustment}: If no: does the eval attempt to adjust for the differences?
        \newline \textit{Select one of: ``Yes'', ``Partial'', \ul{``No'' (Default)}, ``Unknown/Unreported'', or ``N/A''}
    \end{enumerate}
    
    \item \textbf{Tool Access Equivalence}: Did the human baseliners and AIs have access to the same (technical) tools for each item? Respond yes if neither group had access to external tools; respond yes if the human had internet access and the AI did not (but was trained on the internet)
    \newline \textit{Select one of: \ul{``Yes'' (Default)}, ``Partial'', ``No'', ``Unknown/Unreported'', or ``N/A''}
    \begin{enumerate}
        \item \textbf{Tool Access Equivalence Enforcement}: If human baseliners' tool access was limited: was there an oversight mechanism for ensuring that the human baseliners only used the tools permitted? E.g., enforcement of AI tool use ban
        \newline \textit{Select one of: ``Yes'', ``Partial'', \ul{``No'' (Default)}, ``Unknown/Unreported'', or ``N/A''}
    \end{enumerate}
    
    \item \textbf{Explanations}: Did the eval/baseline collect explanations from the human baseliners, after the evaluation was conducted? I.e., explanations for why the human participants responded the way they did
    \newline \textit{Select one of: ``Yes'', ``Partial'', \ul{``No'' (Default)}, ``Unknown/Unreported'', or ``N/A''}
\end{enumerate}

\subsection{Baseline Analysis}

\renewcommand{\labelenumi}{4.\arabic{enumi}}
\renewcommand{\labelenumii}{4.\arabic{enumi}.\arabic{enumii}}
\renewcommand{\labelenumiii}{4.\arabic{enumi}.\arabic{enumii}.\arabic{enumiii}}

\begin{enumerate}[leftmargin=30pt, topsep=0pt, itemsep=0pt]
    \item \textbf{Statistical Significance}: Did the eval test for statistically significant differences between AI and human performance?
    \newline \textit{Select one of: ``Yes'', ``Partial'', \ul{``No'' (Default)}, ``Unknown/Unreported'', or ``N/A''}
    \begin{enumerate}
        \item \textbf{Statistical Significance Test}: If yes: what statistical test was used?
    \end{enumerate}
    
    \item \textbf{Uncertainty Estimate}: Did the paper present a measure of uncertainty for the AI and human baseline results? E.g., confidence intervals, variance, pooled/clustered standard errors, etc.?
    \newline \textit{Select one of: ``Yes'', ``Partial'', \ul{``No'' (Default)}, ``Unknown/Unreported'', or ``N/A''}
    \begin{enumerate}
        \item \textbf{Estimate Type}: Is the reported baseline a point estimate, an interval estimate, or a distribution?
        \newline \textit{Select all that apply}
        \begin{itemize}
            \item \ul{Point estimate (Default)}
            \item Interval estimate
            \item Distribution estimate
        \end{itemize}
    \end{enumerate}
    
    \item \textbf{Evaluation Metric Equivalence}: Was the same evaluation metric measured/compared for both humans and AIs? Respond ``no'' if, e.g., the human baseline is majority vote but the AI baseline is not
    \newline \textit{Select one of: \ul{``Yes'' (Default)}, ``Partial'', ``No'', ``Unknown/Unreported'', or ``N/A''}
    
    \item \textbf{Evaluation Scoring Criteria Equivalence}: Was the same scoring rubric used for both AI and human results?
    \newline \textit{Select one of: \ul{``Yes'' (Default)}, ``Partial'', ``No'', ``Unknown/Unreported'', or ``N/A''}
    
    \item \textbf{Evaluation Scoring Method Equivalence}: Was the same scoring method used for both AI and human results? E.g., human grading, LLM as a judge
    \newline \textit{Select one of: \ul{``Yes'' (Default)}, ``Partial'', ``No'', ``Unknown/Unreported'', or ``N/A''}
    
    \item \textbf{Quality Control Robustness}: If quality controls were implemented: are analyses robust to different choices of exclusion criteria? E.g., do the authors state that the results don't change when including/excluding incomplete data?
    \newline \textit{Select one of: ``Yes'', ``Partial'', ``No'', ``Unknown/Unreported'', or ``N/A''}
\end{enumerate}

\subsection{Baseline Documentation}

\renewcommand{\labelenumi}{5.\arabic{enumi}}
\renewcommand{\labelenumii}{5.\arabic{enumi}.\arabic{enumii}}
\renewcommand{\labelenumiii}{5.\arabic{enumi}.\arabic{enumii}.\arabic{enumiii}}

\begin{enumerate}[leftmargin=30pt, topsep=0pt, itemsep=0pt]
    \item \textbf{Additional Reporting}: Were the following reported?
    \begin{enumerate}
        \item \textbf{Reporting Sample Demographics}: Demographics for human baseliners, e.g., race, gender, etc. Respond yes only if within-sample demographics are reported; e.g., respond no if the paper only reports that 100\% of the sample is based in the U.S.
        \newline \textit{Select one of: ``Yes'', ``Partial'', \ul{``No'' (Default)}, ``Unknown/Unreported'', or ``N/A''}
        
        \item \textbf{Reporting Baseline Instructions}: Instructions/guidelines given to human baseliners
        \newline \textit{Select one of: ``Yes'', ``Partial'', \ul{``No'' (Default)}, ``Unknown/Unreported'', or ``N/A''}
        
        \item \textbf{Reporting Time to Completion}: Time to completion for the eval items
        \newline \textit{Select one of: ``Yes'', ``Partial'', \ul{``No'' (Default)}, ``Unknown/Unreported'', or ``N/A''}
        
        \item \textbf{AI Tool Versions}: AI tools and versions (if baseliners had AI access)
        \newline \textit{Select one of: ``Yes'', ``Partial'', \ul{``No'' (Default)}, ``Unknown/Unreported'', or ``N/A''}

        \item \textbf{Completion Rate}: How many human baseliners were recruited but did not complete the tasks?
        \newline \textit{Select one of: ``Yes'', ``Partial'', \ul{``No'' (Default)}, ``Unknown/Unreported'', or ``N/A''}
    \end{enumerate}
    
    \item \textbf{Baseline Data Availability}: Is the (anonymized) human baseline data publicly available?
    \newline \textit{Select one of: ``Yes'', ``Partial'', \ul{``No'' (Default)}, ``Unknown/Unreported'', or ``N/A''}
    \begin{enumerate}
        \item \textbf{Individual Baseline Data Availability}: If yes: is data available at the individual baseliner level? I.e., can you tell from the dataset which baseliners were responsible for which questions?
        \newline \textit{Select one of: ``Yes'', ``Partial'', \ul{``No'' (Default)}, ``Unknown/Unreported'', or ``N/A''}
        
        \item \textbf{Baseline Data Non-Availability Justification}: If no: is there a reasonable justification for non-disclosure of the baseline dataset? E.g., privacy concerns, safety/security concerns, company policy, etc.
    \end{enumerate}
    \item \textbf{Experimental Materials Availability}: Are experimental materials used to implement the eval/baseline publicly available?
    \newline \textit{Select one of: ``Yes'', ``Partial'', \ul{``No'' (Default)}, ``Unknown/Unreported'', or ``N/A''}
    
    \item \textbf{Analysis Code Availability}: Is the code used to analyze the eval/baseline publicly available?
    \newline \textit{Select one of: ``Yes'', ``Partial'', \ul{``No'' (Default)}, ``Unknown/Unreported'', or ``N/A''}
\end{enumerate}

\clearpage

\section{Appendix: Case Studies} \label{sec:Appendix_Case_Studies}

This appendix contains both positive and negative examples of human baselines. Positive examples of (high quality and highly transparent) human baselines include \citet{wijk_re-bench_2024, legris_h-arc_2024, rein_hcast_2025, brodeur_superhuman_2025}, and we discuss the first two of these below.\footnote{METR has produced a number of high-quality baselines, most recently \citet{rein_hcast_2025} (which was published after our review and therefore not included in our results above). \citet{brodeur_superhuman_2025} was similarly not included as it was not caught by our search terms in the literature review. Our choice to discuss only two examples is not based on the quality of the underlying baselines but due to the fact that we believe the chosen examples to be somewhat more well-known in the ML research community.} We do not claim that positive examples are entirely positive or that they follow all our recommendations in Section \ref{sec:Framework}, merely that they are substantially more robust and transparent than other literature.

For additional negative case studies, see \citet{tedeschi_whats_2023}.

\subsection{Positive Example: \texorpdfstring{\citet{wijk_re-bench_2024}}{Wijk et al. (2024)}}

\textsc{RE-Bench} \citet{wijk_re-bench_2024} is an evaluation of AI research engineering capabilities, and it includes an expert human baseline. We discuss each baseline lifecycle stage below.

\textbf{Design \& Implementation.} The benchmark consists of 7 tasks, on which both humans and AI systems were evaluated. It is unclear if the tasks were iteratively designed, but the authors report validating the tasks on human performance, and they also report significant qualitative results (task trajectories). The sample size was only 61, and no power analysis was conducted, but given the small size of the population of interest\footnote{In fact, the exact size of the population is likely not known with substantial precision.}, it's possible that this sample size is sufficient to ensure reasonable statistical power. The authors do not report whether the study underwent ethics review.

\textbf{Recruitment.} The population of interest is specified as human experts with AI research engineering expertise, i.e., as defined by years of experience, research output, hiring screens, and graduate education. The baseliner sample was recruited using convenience sampling, and the authors used well-defined criteria and rigorous screening methods (including some who completed a CodeSignal screen) as quality controls during the recruitment process. Baseliners were compensated at competitive rates (\$1855 per expert on average).

\textbf{Execution.} It is unclear if quality controls were used post-data collection. The authors somewhat control for method effects by controlling the baseliners' coding environment, and both humans and AI systems were permitted internet access. Level of effort is specified by comparing performance in a specified time interval; all humans were given 8 hours per task, and the authors also report performance when both humans and AI systems are given 2 hours per task. The authors collect logs and other qualitative data.

\textbf{Analysis.} The authors report performance intervals over time, and evaluation metrics, rubrics, and scoring methods are consistent.

\textbf{Documentation.} Significant detail about the baselining methods and the baseliner sample are reported (e.g., professional backgrounds). All task environments are released on Github at \url{https://github.com/METR/ai-rd-tasks/tree/main}, and agent trajectories are also provided at \url{https://transcripts.metr.org/}. The manuscript notes that ``[a]nalysis code and anonymized human expert data [are] coming soon,'' though we were unable to find these materials as of the time of this writing (June 2025).

\subsection{Positive Example: \texorpdfstring{\citet{legris_h-arc_2024}}{LeGris et al. (2024)}}

\textsc{H-ARC} (``Human-ARC'') \cite{legris_h-arc_2024}\footnote{Note that 3 of 4 authors are affiliated with the Department of Psychology of New York University. We notice generally that many high-quality human baselines are created by experts with interdisciplinary backgrounds beyond pure ML research, likely due to more robust scientific research norms in other disciplines. Cf. discussion in \cite{burden_paradigms_2025}.} is the generalist (non-expert) human baseline for \textsc{ARC}, a visual program synthesis benchmark. We discuss each baseline lifecycle stage below.

\textbf{Design \& Implementation.} Human results are collected on the entirety of the \textsc{ARC} evaluation and test sets, ensuring that the test set is consistent across humans and AI systems. It is unclear whether baseline instruments were iterated on. No power analysis was conducted, but the sample size consisted of 1768 baseliners, which is higher than the rule-of-thumb 1000 needed to represent the U.S. adult population (though it is unclear what the exact population of interest is) \cite{gelman_how_2004}.\footnote{Note, though, that the baseliners are unevenly distributed across 800 tasks.} No IRB approval was reported.

\textbf{Recruitment.} No population of interest is explicitly specified, though we can assume the intended population is a large population of human adults (participants were specified to be 18--77 years old). Baseliners are selected via a crowdsourced sample from Amazon Mechanical Turk and CloudResearch. No detailed quality controls are specified, but the CloudResearch service contains built-in tools to improve data quality. Baseliners were compensated \$10 and also awarded a performance bonus of \$1.

\textbf{Execution.} The authors report that some baseliner data is incomplete and conduct a robustness check when excluding and imputing missing data. Baseliners were compensated and also given three tries at each task, which is some measure of effort. The authors collected explanations from baseliners.

\textbf{Analysis.} The authors report interval estimates; no statistical tests were conducted that compared human vs. AI performance, though the authors reported a number of within-sample statistical tests (i.e., comparing different segments of baseliners). Evaluation rubrics and scoring methods are consistent.

\textbf{Documentation.} Significant detail about the baselining methods and the baseliner sample are reported (e.g., age, gender, and other demographic information).\footnote{In fact, \textsc{H-ARC} is released as a separate and independent document from the original benchmark.} Data and code are released on Github at \url{https://github.com/Le-Gris/h-arc}.

\subsection{Positive Examples: Limiting Baseline Interpretations}

Not all human baselines are able to maximize scientific robustness, e.g., due to cost considerations. In these cases, researchers can consider scaling back interpretations of human baselines and clearly outlining methodological shortcomings.

One example of baselines in this vein are those contained in \citet{laine_me_2024}, which is a benchmark of AI systems' levels of situational awareness.\footnote{In fact, the human baselines in \citet{laine_me_2024} are quite thoughtfully designed, and our choice to discuss this baseline is primarily due to the well-justified and well-discussed \textit{interpretations} of the baselines in this paper (not because we believe it to fail any particular robustness criteria).} The authors are careful to conduct the baseline only for relevant tasks, and the baseline is interpreted as an upper-bound on performance.\footnote{More precisely, ``[t]he intended interpretation is that the upper baseline is achievable
and represents a high level of situational awareness (roughly comparable to the role-playing humans).''}

\subsection{Negative Example: \texorpdfstring{\citet{sourati_arn_2024}}{Sourati et al. (2024)}}

\textsc{ARN} \citet{sourati_arn_2024} is a benchmark on ``Analogical Reasoning on Narratives'' and contains a human baseline. We discuss each baseline lifecycle stage below.

\textbf{Design \& Implementation.} Baseliners complete only 120 of 1,096 items, and performance metrics on that subset are then compared to AI performance on the entire evaluation dataset. It is unknown whether baseline instruments were iteratively developed. The population of interest is not specified, but the baseline seems to be intended as a generalist (non-expert) baseline, so a sample size of 2 is clearly insufficient to robustly estimate performance metrics of a broad human population. No IRB approval is reported.

\textbf{Recruitment} No population of interest is specified, and the human baseline was conducted by two research assistants (it is unclear if the baseliners had already been exposed to the evaluation items). No quality controls in recruitment are specified. 

\textbf{Execution} Almost nothing about the baseline execution is specified, though the authors do provide the instructions given to baseliners.

\textbf{Analysis} Only point estimates are reported. Furthermore, the language used to report human performance somewhat overstates baseline results. For instance, it is difficult to make claims that ``models are not as good as humans at distinguishing analogies from distractors (57 vs 96\%)'' when there are only two baseliners.\footnote{A more accurate statement might be along the lines of ``models are not as good as two undergraduate students at [university] . . . .''}

\textbf{Documentation} The authors do not report most methodological details, and all that is known about the baseliner sample is that it consists of two research assistants. As of this writing, we are unable to find publicly available code or baseline results (though the benchmark items are available online). 

\subsection{Negative Examples: Non-Transparent Reporting}

A number of studies report almost nothing about their human baselines except for the results. Some examples include: \citet{chiu_culturalbench_2024, jing_followeval_2023, mukhopadhyay_mapwise_2024, yue_mmmu_2024, blinov_rumedbench_2022, gong_bloomvqa_2024, yin_automl_2024}. For instance, \citet{gong_bloomvqa_2024} contains only two sentences about its human baseline:

\begin{quote}
    ``We perform human baseline on proposed dataset (1200 core VQA samples) with a small group of adult reviewers. The human baseline reports an average accuracy of 89\% with 2\% standard deviation.''
\end{quote}

It is nearly impossible to interpret this result: how many ``adult reviewers'' participated in the baseline? How were these baseliners selected? What instructions were given to these baseliners? Without substantially more detail about study methodology and about the baseliner sample, readers are entirely unable to determine the quality of the baseline and the extent to which the baseline is generalizable to other populations or settings.

\clearpage

\section{Appendix: Discussion on Expert Human Baselines} \label{sec:Appendix_Expert_Baselines}

Given the rapid advancement of AI systems, researchers, policymakers, and the general public have an interest in benchmarking AI capabilities against those of the highest performing humans in a given domain. As such, many human baselines are now expert baselines. Some preliminary discussion of these baselines follows:

\begin{itemize}
    \item Expert baselines are well-suited for estimating the \textit{maximum} possible human performance---i.e., the best that humans can currently perform on an evaluation item (as in, e.g., \citealt{obeidat_arentail_2024, laine_me_2024}).\footnote{Researchers could also attempt to estimate the mean of expert performance, but this estimation would be more difficult as expert populations are often unknown size, smaller than, and less accessible than non-expert populations.}
    \item For any expert baselines, evaluators should develop and report clear standards for what constitutes expertise. Note that not all expertise is tied to professional experience or educational credentials. See \citet{diaz_what_2024} for a discussion on constructions of ``expertise'' in machine learning research.
    \item Expert populations are often small, and smaller baseliner sample sizes are increasingly more acceptable as evaluation items become more specialized. 
    \item Given the nature and often small population of human experts within a given domain, evaluators may collect convenience samples while maintaining clear standards on expertise. See \citet{wijk_re-bench_2024} for an example of rigorous recruitment criteria in a convenience sample. Evaluators can also consider snowball sampling---i.e., sampling by asking study participants to recruit other participants from their networks into the study; see \citet{parker_snowball_2019} for an overview of snowball sampling.
    \item Another example of an expert human baseline is \citet{asiedu_contextual_2025}.
    \item When calculating results, taking the maximum of all scores per item is acceptable for estimating maximum performance. Note, however, that sample maxima have extreme distributions, and measures of uncertainty should be calculated differently compared to other sample statistics.
    \item As foundation models are trained on data from across the internet (and thus have ``seen'' relevant information already), comparing expert performance with internet access with AI system performance is likely a fair comparison, even where the AI system does not have internet access.
\end{itemize}

\clearpage

\section{Appendix: Methodology} 
\label{sec:Appendix_Methods}

We adopted a two-stage methodology as described in Section \ref{sec:Methods}, adapted from the methodology of \citet{zhao_position_2025} and \citet{reuel_betterbench_2024}. 

Section \ref{subsec:Appendix_Meta-Review} describes stage one, in which we conducted a meta-review of the measurement theory and AI evaluation literatures to qualitatively synthesize the checklist in Appendix \ref{sec:Appendix_Checklist}. 

Section \ref{subsec:Appendix_Lit_Review} describes stage two, in which we systematically reviewed human baselines in foundation model evaluations. 

\subsection{Meta-Review}
\label{subsec:Appendix_Meta-Review}

We begin with a scoping meta-review (a review of reviews) to qualitatively identify and synthesize literature relevant to human baselining. Meta-reviews are useful when there is little direct literature on the research question of interest (here, human baselines) but there is relevant literature from related fields (here, measurement theory) \cite{sarrami-foroushani_scoping_2015}. As there is a wealth of literature in measurement theory, a meta-review that synthesizes the relevant evidence is appropriate to collect evidence in one place and to prevent researchers from being overwhelmed by the quantity of evidence \cite{hennessy_best_2019}.

Our literature search process adopted a purposive sampling approach. Although a systematic search process is normally ideal \cite{hennessy_best_2019}, purposive sampling is also acceptable for qualitative literature synthesis \citetext{e.g., \citealt{ames_purposive_2019}} and is justified here due to the broad scope of the relevant literature \cite{palinkas_purposeful_2015}. Our sampling approach used theory-based inclusion criteria \cite{palinkas_purposeful_2015}: we queried Google Scholar and \citet{annual_reviews_annual_2025} in December 2024 for the keywords in Table \ref{tab:Metareview_Inclusion}, then filtered according to the criteria in Table \ref{tab:Metareview_Inclusion}. We also conducted backwards snowballing for the ML articles to identify further relevant literature. Finally, we added items to the sample based on our expertise, as many of the authors have experience in social science methodology and AI evaluation.

One limitation of this search strategy is that it introduces some sampling bias due to searching directly on the Annual Review website. We consider this limitation acceptable because by impact factor, Annual Reviews is a top-ranked publisher of literature reviews in the relevant social science disciplines (e.g., political science, psychology, sociology, statistics, economics) \cite{annual_reviews_journal_2025}. We thus expect our meta-review sample to be high-quality and relatively high-coverage.

\begin{table}[!htbp]
    \centering
    \begin{tabularx}{\linewidth}{ X X }
        \toprule
        
        \textbf{Type} &
        \textbf{Inclusion Criteria}
        \\ 
        
        \midrule

        Document type &
        Literature review
        \newline Position paper
        \newline Synthesis article
        \newline Book or book chapter (including reference texts)
        \vspace{0.4em}
        \\

        Subject area &
        Measurement theory (including applications in statistics, economics, political science, psychology, education, sociology, or medicine)
        \newline AI evaluation
        \vspace{0.4em}
        \\

        Keywords (non-exhaustive) &
        ``measurement theory''
        \newline ``measurement model*''
        \newline ``validity''
        \newline ``reliability''
        \newline ``replicability''
        \newline ``survey design''
        \newline ``survey method*''
        \newline ``questionnaire design''
        \newline ``experimental design''
        \newline ``causal inference''
        \\
        
        \bottomrule
    \end{tabularx}
\caption{Inclusion criteria for meta-review articles.}
\label{tab:Metareview_Inclusion}
\end{table}

Our search process yielded a total of 29 articles to be included in our meta-review (listed in Table \ref{tab:Metareview_Articles}). To synthesize our checklist, KW scanned these 29 articles and compiled a list of relevant methodological practices/considerations in a Google Sheet, categorizing each into the categories of baseline(r) design, recruitment, execution, analysis, and documentation. The authors then collectively discussed the checklist and validated the checklist using expert feedback from six external experts before refining and finalizing the checklist. Finally, the checklist was also iteratively refined during the coding process.

\begin{table*}[!htbp]
    \centering
    \begin{tabularx}{\textwidth}{l X}
        \toprule

        \textbf{Subject area} &
        \textbf{Articles}
        \\

        \midrule

        Measurement theory ($n = 17$) &
        \citet{bandalos_measurement_2018, berinsky_measuring_2017, cai_item_2016, chang_statistical_2021, couper_new_2017, findley_external_2021, groves_survey_2011, imbens_causal_2015, jackson_principles_2013, kertzer_experiments_2022, list_so_2011, nosek_replicability_2022, rosellini_developing_2021, stantcheva_how_2023, strauss_construct_2009, zhang_statistical_2023, zickar_measurement_2020}
        \vspace{0.4em}
        \\
        
        Machine Learning ($n = 12$) &
        \citet{agarwal_deep_2022, bowman_what_2021, cowley_framework_2022, dow_dimensions_2024, eckman_position_2025, ibrahim_beyond_2024, liao_are_2021, reuel_betterbench_2024, subramonian_it_2023, wang_evaluating_2023, xiao_evaluating_2023, zhou_deconstructing_2022}
        \\

        \bottomrule
    \end{tabularx}
    \caption{A complete list of the 29 articles included in our meta-review.}
    \label{tab:Metareview_Articles}
\end{table*}

\subsection{Systematic Literature Review}
\label{subsec:Appendix_Lit_Review}

We conducted a systematic literature review of human baselines in AI evaluations \cite{page_prisma_2021} to identify gaps in baselining methodology. Our review method is similar to that of \citet{zhao_position_2025}.

First, we conducted a systematic search for relevant literature. To begin, we queried Google Scholar in December 2024 for articles containing the keywords in Table \ref{tab:Review_Keywords}.\footnote{Note on terminology: the literature currently has no unified, standard terminology for referring to human baselines. Terms such as ``human baseline,'' ``expert baseline,'' and ``human performance baseline'' (the terms contained in Table \ref{tab:Review_Keywords}) are commonly used; however, some authors use ``human benchmark'' or even ``human evaluation.'' We discovered usage of the term ``human benchmark'' after our search was concluded; although a qualitative analysis indicated that this term was less frequently used than those in ``human benchmark,'' and a spot check leads us to believe that our results would not differ significantly even with inclusion of this term, the exclusion of this search term is nevertheless a limitation to the coverage of our systematic search. We deliberately excluded ``human evaluation'' from our search terms, as this term is normally used to describe human annotations, grading, or scoring of AI outputs rather than human baselines \citetext{e.g., as used in \citealt{howcroft_twenty_2020}}, though a number of our included articles nevertheless contained this term.} Our search terms were intentionally broad, as authors use a variety of different language to describe human baselines. Articles were included in the initial sample if they contained in the full text both a human baseline keyword and an AI evaluation keyword. 

\begin{table}[!htb]
    \centering
    \begin{tabularx}{\linewidth}{ X X }
        \toprule
        \textbf{Type} & 
        \textbf{Keywords} 
        \\
        
        \midrule

        Human Baseline Keywords & 
        ``human baseline*'' \newline ``expert baseline*'' \newline ``human performance baseline*''
        \vspace{0.4em}
        \\

        AI Evaluation Keywords & 
        ``LLM evaluation*'' \newline ``AI evaluation*'' \newline ``NLP evaluation*'' \newline ``ML evaluation*'' \newline ``model evaluation*'' \newline ``LLM benchmark*'' \newline ``AI benchmark*'' \newline ``NLP benchmark*'' \newline ``ML benchmark*'' \newline ``evaluating LLM*'' \newline ``evaluation of LLM*'' \newline ``benchmark LLM'' \newline ``benchmarking LLMs'' \newline ``evaluation of AI models''
        \\
        
        \bottomrule
    \end{tabularx}
    \caption{Search terms for systematic literature review of human baselines}
    \label{tab:Review_Keywords}
\end{table}

Google Scholar was chosen as the database of choice due to its comprehensive coverage \cite{gusenbauer_google_2019} and its indexing of the gray literature. We included articles in the gray literature (e.g., preprints) because researchers often post preprints on arXiv prior to formal publication and because a substantial portion of ML literature is published on arXiv, including publications from many industry organizations \cite{shah_jahan_bidirectional_2021}. For instance, arXiv was the source of an overwhelming majority of articles in one recent systematic literature review on bidirectional language models \cite{shah_jahan_bidirectional_2021}. 

There is debate in the methodological literature about the use of Google Scholar as a primary database in a systematic literature review. Concerns have been raised about limitations to advanced search capabilities and to the Google Scholar interface \cite{halevi_suitability_2017}, lack of precision \cite{boeker_google_2013}, and lack of coverage \cite{haddaway_role_2015}. We addressed these limitations as follows:

\begin{itemize}[itemsep=0pt, topsep=0pt]
    \item To address limitations to advanced search capabilities, we did not use advanced search capabilities beyond the boolean AND and OR operators in search strings, as well as a simple date filter. 
    \item To address interface limitations, we created workarounds by using multiple queries (to avoid the 256 character limit in search strings) and using a bookmarklet to capture reference information. In any case, we generally find that the search capabilities and interface of Google Scholar are an improvement over the search function in arXiv, making queries to Google Scholar preferable to direct queries in arXiv. 
    \item To address limitations in precision, we adopted more stringent inclusion/exclusion criteria to filter our sample (discussed below). Furthermore, our search is necessarily imprecise due to a lack of standardization of terms in describing human baselines in the literature (e.g., we found that some literature described baselines as ``human evaluation'', which is normally used to describe human annotations of evaluation data).
    \item To address limitations in coverage, we supplemented our Google Scholar search with other sources. SD queried Elicit\footnote{Elicit is an AI search and analysis for researchers \cite{elicit_elicit_nodate}.} for articles containing human baselines, and MB identified evaluation datasets with human baselines used in industry evaluations by scanning the model cards/system cards of OpenAI o1 \cite{openai_openai_2024}, Anthropic's Claude 3.5 Sonnet \cite{anthropic_claude_2024}, Meta's Llama 3 \cite{grattafiori_llama_2024}, and Google DeepMind's Gemini 1.5 \cite{gemini_team_google_gemini_2024}.\footnote{The date filter in Table \ref{tab:Review_Exclusion} was not applied for these articles so that we could capture evaluations that are widely used in practice. Only one article that would have otherwise been excluded was ultimately included in our sample of baselines \cite{dua_drop_2019}.} Furthermore, the most recent research has found that Google Scholar has significantly expanded its coverage \cite{gusenbauer_google_2019}, and another study found that Google Scholar indexed 96\% of articles in systematic literature reviews in computer science that were conducted using other databases \cite{yasin_using_2020}. 
\end{itemize}

Our search process yielded a sample of $n = 397$ articles (378 from Google Scholar, 13 from Elicit, and 6 from industry model/system cards), which were stored in a Google Sheet. KW then scanned the title, abstract, and main text of each article to filter the sample; the inclusion/exclusion criteria used in filtering along with rationales for each criterion are discussed in Tables \ref{tab:Review_Inclusion} and \ref{tab:Review_Exclusion}.\footnote{Note that although one of our exclusion criteria is for articles published before 2020, we make one exception and nevertheless include one article from 2019 \cite{dua_drop_2019} due to its prevalence in industry model cards.} As Google Scholar does not always index the most authoritative version of articles, KW also cross-referenced DBLP for all articles on preprint servers (including arXiv) to identify the latest version or published version of each preprint.\footnote{Since the time of this writing, published versions of four articles we reviewed have become available. We cite to the preprints that we reviewed below \cite{de_haan_astromlab_2024, bai_power_2024, mukhopadhyay_mapwise_2024, lei_iwisdm_2024}, but the published versions are available at: \citealt{de_haan_achieving_2025, bai_power_2025, mukhopadhyay_mapwise_2025, lei_iwisdm_2025}.} During the coding process, all coders were also made aware of the exclusion criteria in case any invalid articles were inadvertently included for coding. The final number of articles included for analysis was $n = 109$, and these are identified in Table \ref{tab:Review_Result_Articles}. 

\begin{table*}[!htbp]
    \centering
    \begin{tabularx}{\linewidth}{ X X }
        \toprule
        
        \textbf{Inclusion Criteria} &
        \textbf{Rationale}
        \\ 

        \midrule
        
        Article contains an evaluation of a foundation model &
        \begin{itemize}[after=\vspace{-\baselineskip}, topsep=0pt, itemsep=0pt, parsep=0pt]
            \vspace{-0.75em}
            \item We limited our scope to foundation models in part to make the review practically manageable
            \item No comprehensive guidance exists for human baselines that is specific to the context of foundation models and that accounts for the most recent foundation model literature
            \item Foundation models raise different and somewhat unique considerations for human baselines, and we aimed to narrow in on these specific considerations
            \item Examples of qualifying articles: articles that fine-tuned or used pre-trained large (language or multi-modal) models
            \vspace{0.4em}
        \end{itemize}
        \\

        Article contains a human baseline (defined in Section \ref{sec:Intro}) &
        \begin{itemize}[after=\vspace{-\baselineskip}, topsep=0pt, itemsep=0pt, parsep=0pt]
            \vspace{-0.75em}
            \item See exclusion criteria for examples of non-qualifying articles
            \vspace{0.4em}
        \end{itemize}
        \\

        Article is published in a peer-reviewed venue or is available in the gray literature (e.g., on a preprint server such as arXiv) &
        \begin{itemize}[after=\vspace{-\baselineskip}, topsep=0pt, itemsep=0pt, parsep=0pt]
            \vspace{-0.75em}
            \item See text for a discussion of arXiv
        \end{itemize}
        \\
        
        \bottomrule
    \end{tabularx}
\caption{Inclusion criteria for systematic review of human baselines.}
\label{tab:Review_Inclusion}
\end{table*}

\begin{table*}[!htbp]
    \centering
    \begin{tabularx}{\linewidth}{ X X }
        \toprule
        
        \textbf{Exclusion Criteria} &
        \textbf{Rationale}
        \\ 

        \midrule
        
        AI model being evaluated is not a foundation model &
        \begin{itemize}[after=\vspace{-\baselineskip}, topsep=0pt, itemsep=0pt, parsep=0pt]
            \vspace{-0.75em}
            \item See inclusion criteria for discussion of rationale
            \item Examples of non-qualifying articles: articles that trained non-general purpose models for specific purposes
            \vspace{0.4em}
        \end{itemize}
        \\

        Article did not contain a human baseline &
        \begin{itemize}[after=\vspace{-\baselineskip}, topsep=0pt, itemsep=0pt, parsep=0pt]
            \vspace{-0.75em}
            \item Enforcement of analogous inclusion criteria
            \vspace{0.4em}
        \end{itemize}
        \\
        
        Human baseline in the article is not original (e.g., uses observational/real-world data or a human baseline from a pre-existing dataset) &
        \begin{itemize}[after=\vspace{-\baselineskip}, topsep=0pt, itemsep=0pt, parsep=0pt]
            \vspace{-0.75em}
            \item Articles using observational/real-world data are excluded as it is difficult to make direct comparisons between human and AI performance in such cases, given that the human data was not generated in a controlled laboratory setting
            \item Articles using pre-existing human baseline data are excluded as researchers may fail to adhere to the experimental design of the previous baseline, making comparisons difficult
            \vspace{0.4em}
        \end{itemize}
        \\
        
        Article duplicates an item already included in the review &
        \begin{itemize}[after=\vspace{-\baselineskip}, topsep=0pt, itemsep=0pt, parsep=0pt]
            \vspace{-0.75em}
            \item Prevention of duplicate items
            \item Examples of non-qualifying items: preprint or workshop version of subsequently published work
            \vspace{0.4em}
        \end{itemize}
        \\
        
        Article was published before 2020 &
        \begin{itemize}[after=\vspace{-\baselineskip}, topsep=0pt, itemsep=0pt, parsep=0pt]
            \vspace{-0.75em}
            \item Most foundation model evaluation literature was published after 2020 (inclusive)
            \vspace{0.4em}
        \end{itemize}
        \\
        
        Article collected data from human annotators but not as a baseline &
        \begin{itemize}[after=\vspace{-\baselineskip}, topsep=0pt, itemsep=0pt, parsep=0pt]
            \vspace{-0.75em}
            \item Use of human data in non-baseline contexts gives rise to different methodological considerations
            \item Examples of non-qualifying articles: articles using human evaluation (i.e., using human annotators to score or analyze evaluation data), articles collecting human data as ground truth (e.g., using annotations to determine the desired responses to evaluation items)
            \vspace{0.4em}
        \end{itemize}
        \\
        
        Article evaluates LLM-as-a-judge, i.e., compares LLM vs. human evaluation of AI models &
        \begin{itemize}[after=\vspace{-\baselineskip}, topsep=0pt, itemsep=0pt, parsep=0pt]
            \vspace{-0.75em}
            \item LLM-as-a-judge may give rise to highly idiosyncratic methodological considerations 
            \vspace{0.4em}
        \end{itemize}
        \\
        
        Article is incomplete work or work-in-progress &
        \begin{itemize}[after=\vspace{-\baselineskip}, topsep=0pt, itemsep=0pt, parsep=0pt]
            \vspace{-0.75em}
            \item Quality control
            \item Examples of non-qualifying articles: articles submitted to venues but not released as preprints (e.g., paper available on OpenReview but not on arXiv; we assume that authors of these articles do not intend to make their papers public), articles submitted to non-archival workshops intended to refine work
            \vspace{0.4em}
        \end{itemize}
        \\
        
        Article is a thesis or class work &
        \begin{itemize}[after=\vspace{-\baselineskip}, topsep=0pt, itemsep=0pt, parsep=0pt]
            \vspace{-0.75em}
            \item Quality control
        \end{itemize}
        \\
        
        \bottomrule
    \end{tabularx}
\caption{Exclusion criteria for systematic review of human baselines.}
\label{tab:Review_Exclusion}
\end{table*}

Following the coding strategy in \citet{zhao_position_2025}, a subset of authors each coded the same four articles, discussed results to ensure coding consistency, and refined the checklist items. The remaining articles were then split up for coding between all authors, with results stored in a Google Sheet. Questions that arose during the final coding process were adjudicated via discussion. All codes were then validated by KW, PP, SD, and MB (with no coder validating their own codes). After validation was complete, KW cleaned and standardized the final dataset.

Note that many articles conducted human baselines for multiple different datasets. During the coding process, each dataset for which a baseline was conducted was coded separately. During the process of cleaning and analyzing coded data, only baselines that contained key differences in baseline instrument, construct, sample, or process were retained as distinct baselines; these were formalized as different coding for Q1.5--1.8, Q2.1--2.3, Q2.6--2.7, or Q3.2--3.5. We find six articles which contained more than one distinct baselines by this criteria \cite{lu_newsinterview_2024, meister_benchmarking_2024, castro_fiber_2022, verma_ghostbuster_2024, laine_me_2024, suvarna_phonologybench_2024}, bringing the total number of human baselines up to 115.

\begin{table*}[!htbp]
    \centering
    \begin{tabularx}{\textwidth}{l X}
        \toprule

        &
        \textbf{Articles}
        \\

        \midrule

        Included ($n = 109$) & 
        \citet{abdibayev_bpomp_2021, akhtar_chartcheck_2024, albrecht_despite_2022, alex_raft_2021, asami_propres_2023, asiedu_contextual_2025, awal_vismin_2025, bai_power_2024, blinov_rumedbench_2022, bu_roadmap_2024, castro_fiber_2022, chang_partnr_2024, chang_speak_2023, chen_tombench_2024, chiu_culturalbench_2024, chiyah-garcia_repairs_2024, costarelli_gamebench_2024, dagli_airletters_2024, dua_drop_2019, duan_pip_2022, fenogenova_mera_2024, fyffe_transforming_2024, gong_bloomvqa_2024, gu_detectbench_2024, guo_can_2024, gupta_polymath_2024, de_haan_astromlab_2024, hackenburg_comparing_2023, hamotskyi_eval-ua-tion_2024, heiding_evaluating_2024, MATH, hijazi_arablegaleval_2024, hildebrandt_odd-dillmma_2024, hou_entering_2024, huang_apathetic_2024, ivanov_biolp-bench_2024, jain_language_2023, ji_abstract_2022, jimenez_carets_2022, jing_followeval_2023, kodali_human_2024, kruk_silent_2024, lacombe_climatex_2023, laine_me_2024, laurent_lab-bench_2024, legris_h-arc_2024, lei_iwisdm_2024, li_naturalbench_2024, li_search_2024, li_value_2021, li_vitatecs_2025, lin_truthfulqa_2022, liu_learning_2024, liu_webdp_2023, liu_tempcompass_2024, lu_newsinterview_2024, lu_mathvista_2023, mangalam_egoschema_2023, meister_benchmarking_2024, mialon_gaia_2023, miller_effect_2020, mirza_are_2024, mizrahi_coming_2020, montalan_kalahi_2024, moskvichev_conceptarc_2023, mukhopadhyay_mapwise_2024, norlund_transferring_2021, obeidat_arentail_2024, phuong_evaluating_2024, reese_comparing_2024, rein_gpqa_2024, roberts_scifibench_2024, ruis_goldilocks_2023, sakai_mcsqa_2024, santurkar_breeds_2020, sanyal_are_2024, shavrina_russiansuperglue_2020, si_can_2024, someya_jblimp_2023, sourati_arn_2024, sprague_musr_2023, srivastava_beyond_2023, suvarna_phonologybench_2024, tahsin_mayeesha_deep_2021, taktasheva_tape_2022, tanzer_benchmark_2023, thrush_i_2024, valmeekam_planning_2023, verma_ghostbuster_2024, wadhawan_contextual_2024, webson_are_2023, weissweiler_hybrid_2024, wijk_re-bench_2024, wu_visco_2024, wu_smartplay_2023, xiang_care-mi_2023, yin_automl_2024, yue_mmmu_2024, zamecnik_mapping_2024, zerroug_benchmark_2022, zhang_mathemyths_2024, zhang_hire_2024, zhang_probing_2024, zhou_diffusyn_2024, zhou_labsafety_2024, zhu_excalibur_2023, zhuo_bigcodebench_2024}
        \\

        \bottomrule
    \end{tabularx}
    \caption{A complete list of the 109 articles included in our systematic review of human baselines. Note that we analyze 115 individual baselines from these articles, as a single article may contain multiple baselines (see explanation in text).}
    \label{tab:Review_Result_Articles}
\end{table*}

\clearpage

\section{Appendix: Additional Resources} \label{sec:Appendix_Resources}

This appendix contains a non-exhaustive list of further resources for researchers interested in running a human baseline. Most items in this list are cited in the relevant sections above, and we provide this list purely for convenience.

For practical guidance on designing human baselines (and surveys):
\begin{itemize}
    \item Aside from this paper, your first stop for highly practical guidance on baseline design should be the appendix in \citet{stantcheva_how_2023},\footnote{Available directly at: \url{https://www.annualreviews.org/content/journals/10.1146/annurev-economics-091622-010157\#supplementary_data} (archived at \url{https://perma.cc/EZ9X-XK6A}).} which contains detailed information on sampling methods, crowdsourcing, writing survey questions, response bias, and other useful resources. 
    \item For guidance on validity in AI evaluations, see \citet{salaudeen_measurement_2025}.
    \item For guidance on power analysis, see \citet{card_little_2020}.
    \item For best practices on ensuring data quality on Amazon Mechanical Turk, see \citet{lu_improving_2022}.
    \item For methods to prevent AI usage in human baselines, see \citet{veselovsky_prevalence_2023}.
    \item For additional best practices in AI evaluation, see \citet{reuel_betterbench_2024, paskov_gpai_2024, biderman_lessons_2024, grosse-holz_early_2024}.
    \item For a summary on uncertainty estimation and other statistical methods, see \citet{miller_adding_2024} (with the caveat of \citealt{bowyer_position_2025}).
    \item For early work on dealing with small sample sizes in evaluations, see \citet{luettgau_hibayes_2025, xiao_confidence_2025}.
    \item For best practices on reproducibility and transparency in ML research, see \citet{kapoor_reforms_2024, semmelrock_reproducibility_2024}.
\end{itemize}

For more comprehensive reference texts:
\begin{itemize}
    \item For an introduction to measurement theory, see \citet{bandalos_measurement_2018}.
    \item For an introduction to (human) survey research, see \citet{groves_survey_2011} or \citet{valliant_practical_2018}.
    \item For an introduction to survey sampling (likely more complex than is necessary for most evaluations), see \citet{lohr_sampling_2022}.
    \item For an introduction to survey weights, see \citet{valliant_practical_2018}.
    \item For an introduction to hierarchical modelling and sample size estimation, see chapters 8 and 11 of \citet{mcnulty_power_2021} (chosen as it provides examples in Python).
    \item For a discussion of measurement equivalence or measurement invariance (ensuring that measurement instruments capture the same concept across different populations), see \citet{davidov_measurement_2014}.\footnote{Although this discussion is in the context of measurements in different populations of \textit{humans}, many measurement equivalence questions also arise between populations of humans and AI models.}
    \item For discussions on using and interpreting $p$-values in statistics, see \citet{mcshane_abandon_2019, gelman_difference_2006}. 
\end{itemize}

\clearpage

\section{Appendix: Data Availability} \label{sec:Appendix_Data_Availability}

An up-to-date version of our checklist as well as individual annotations from our systematic review of human baselines are available at: \url{https://github.com/kevinlwei/human-baselines}.

\end{document}